# PsychBench: A comprehensive and professional benchmark for evaluating the performance of LLM-assisted psychiatric clinical practice

Shuyu Liu[1†], Ruoxi Wang[1†], Ling Zhang[2,3†], Xuequan Zhu[2,3†], Rui Yang[2,3], Xinzhu Zhou[2], Fei Wu[1], Zhi Yang[2,3], Cheng Jin[1,2,4*], Gang Wang[2,3*]

[1*]School of Biomedical Engineering, Shanghai Jiao Tong University, Shanghai, 200240, China.

[2*]Beijing Key Laboratory of Mental Disorders, National Clinical Research Center for Mental Disorders National Center for Mental Disorders, Beijing Anding Hospital, Capital Medical University, Beijing, 100088, China.

[3*]Advanced Innovation Center for Human Brain Protection, Capital Medical University, Beijing, 100088, China.

[4*]Shanghai Artificial Intelligence Laboratory, Shanghai, 200232, China

[†]These authors contributed equally to this work.

## Abstract

The advent of Large Language Models (LLMs) offers potential solutions to address problems such as shortage of medical resources and low diagnostic consistency in psychiatric clinical practice. Despite this potential, a robust and comprehensive benchmarking framework to assess the efficacy of LLMs in authentic psychiatric clinical environments is absent. This has impeded the advancement of specialized LLMs tailored to psychiatric applications. In response to this gap, by incorporating clinical demands in psychiatry and clinical data, we proposed a benchmarking system, PsychBench, to evaluate the practical performance of LLMs in psychiatric clinical settings. The PsychBench is composed of a comprehensive dataset and an evaluation framework. The dataset includes 300 real-world patient cases sourced from three geographically diverse medical centers across northern, central, and southern China, ensuring broad regional and cultural representation. The evaluation framework encompasses one psychiatric knowledge evaluation task and five key clinical tasks—clinical text understanding and generation, principal diagnosis, differential analysis, medication recommendation, and long-term course management—each supported by psychiatry-specific quantitative evaluation metrics to ensure rigorous performance assessment. We conducted a comprehensive quantitative evaluation of 16 LLMs using PsychBench, and investigated the impact of prompt design, chain-of-thought reasoning, input text length, and domain-specific knowledge fine-tuning on model performance. Through detailed error analysis, we identified strengths and potential limitations of the existing models and suggested directions for improvement. Subsequently, a clinical reader study involving 60 psychiatrists of varying seniority was conducted to further explore the practical benefits of existing LLMs as supportive tools for psychiatrists of varying seniority. Through the quantitative and reader evaluation, we show that while existing models demonstrate significant potential, they are not yet adequate as decision-making tools in psychiatric clinical practice. The reader study further indicates that, as an auxiliary tool, as an auxiliary tool, current LLMs could provide effective support for junior psychiatrists, effectively enhancing their work efficiency and the comprehensiveness of analytical work. To promote research in this area, we will make the dataset and evaluation framework publicly available, with the hope of advancing the application of LLMs in psychiatric clinical settings.

**Keywords**: Mental health, Large language models, Benchmark, Clinical practice

# Introduction

In recent years, the prevalence of mental disorders has been steadily increasing, becoming a major global public health challenge[1,2]. However, this rising number of patients contrasts sharply with the relative scarcity of mental health resources, particularly in terms of the availability of psychiatrists and access to specialized care[3,4]. This imbalance has driven the exploration of new technologies in psychiatric practice. Against this backdrop, the emergence of LLMs presents a new potential solution to this issue. Given the heavy reliance on verbal communication and text analysis in psychiatric care, LLMs demonstrate a greater application advantage in supporting diagnosis, treatment, and patient management in psychiatry compared to other fields focused on organic diseases. By rapidly analyzing and interpreting patients' emotional expressions, thought patterns, and linguistic features, LLMs can offer real-time, intelligent decision support for psychiatrists[5-12]. However, to implement LLMs effectively in psychiatric clinical practice, it is essential to ensure their comprehensive and reliable performance, which necessitates systematic and scientific evaluation. Currently, research on evaluating the performance of LLMs in psychiatric applications is still in its early stages, lacking sufficient empirical evidence and evaluation frameworks. This underscores the urgent need of an evaluation framework to explore and validate the feasibility and effectiveness of LLMs in psychiatric clinical practice.

At present, the evaluation of LLMs mainly revolves around standardized exams and simulated clinical data, where models are presented with straightforward information and multiple-choice options, requiring little in-depth analysis to reach an answer. Multiple studies have shown that LLMs perform exceptionally well on these tests, achieving results comparable to, or even surpassing, those of human doctors in medical knowledge and diagnostic reasoning, suggesting strong capabilities in processing medical information[6,7,13-21]. However, these evaluation methods mainly emphasize static and general knowledge assessment, fails to fully capture the model's response when faced with complex patient scenarios in real-world medical settings[22,23], especially in psychiatry, which requires interpreting multi-stage, longitudinal patient records and integrating evolving information across repeated clinical encounters.

Many studies have attempted to explore the potential and limitations of LLMs in psychiatric health care. One investigation assessed ChatGPT's performance across three simulated psychiatric cases with varying complexity, revealed concerning limitations of LLMs in clinical reasoning, information synthesis, and safety-critical judgment, especially in complex scenarios[24]. Recent efforts have evaluated LLMs in mental health contexts, including psychotherapy simulations[25], proactive conversational coaching[26], motivational interviewing[27], and emotion classification using social media data[28]. Some studies have assessed the performance of LLMs in making clinical decisions for bipolar disorder and offering treatment recommendations for mild depression using hypothetical clinical vignettes[8,10]. In parallel, a recent *Lancet Digital Health* viewpoint emphasized the need for representative datasets, ethical and inclusive deployment strategies, and greater clinical contextualization when applying LLMs in mental health care[29]. While these studies mark important progress, they often rely on simulated interactions, center on limited task types, or focus more on conversational behavior and adherence to design principles rather than clinically grounded decision-making. Rigorous, real-world evaluations and standardized frameworks are still needed to assess the safety and effectiveness of LLMs in psychiatric practice.

To develop a comprehensive, professional, and reliable evaluation system for LLMs in the clinical field of psychiatry, it is essential to ground the design in real-world clinical practice. This requires leveraging authentic medical data, adhering strictly to evidence-based clinical guidelines, and addressing the full diversity and complexity of actual psychiatric care needs[30-32]. Informed by expert consensus from our clinical committee, the key areas where LLMs can assist psychiatric practice can be summarized into five primary tasks. (1) Clinical text understanding and generation. Psychiatrists spend significant time drafting medical records[12]. By automating document generation, LLMs can free up psychiatrists' time, allowing them to allocate more of their day to patient interaction. (2) Principal diagnosis. Diagnosing psychiatric disorders often involves interpreting complex symptoms and subjective descriptions, and psychiatric clinical practitioners have been criticized for not sufficiently adhering to evidence-based clinical guidelines[33]. Accurate and robust diagnostic assistance is crucial for helping psychiatrists analyze patient symptoms effectively and enhancing diagnostic accuracy. (3) Differential analysis. The frequent symptom overlap across psychiatric disorders necessitates comprehensive differential analysis capabilities to minimize diagnostic errors. LLMs should offer comprehensive analysis given patient information, assisting psychiatrists in ruling out misdiagnosis. (4) Medication recommendation. Psychiatric pharmacotherapy requires nuanced consideration of individual patient profiles. Clinicians need help from LLMs to synthesize complex clinical data to generate medication recommendations aligned with evidence-based protocols. (5) Long-term course management. During long-term course management, rapid and reliable information retrieval is necessary and is the key to improving the efficiency[34]. By quickly analyzing historical patient data, LLMs can aid psychiatrists in developing effective long-term treatment plans and providing real-time support during consultations.

An LLM can be considered suitable for integration into psychiatric clinical workflows if it fulfills three core criteria. First, it must demonstrate performance comparable to psychiatrists with intermediate seniority across key clinical functions, providing valuable insights to doctors of all seniority to enhance overall psychiatric care even in low-resource setting. Second, a clinically useful LLM must demonstrate a deep understanding of psychiatric knowledge and apply it rigorously to the thinking and decision-making processes of professional physicians. It should align with existing clinical guidelines and clearly and transparently tailor the output to the individual patient, enabling clinicians to follow and trust its recommendations and gain valuable insights. Third, it should handle long, text-rich medical records with efficiency and fidelity, reducing documentation burdens, enhancing work efficiency without sacrificing clinical nuance.

In this study, a comprehensive evaluation system, PsychBench, was proposed for assessing the performance of LLMs in the clinical psychiatric field. The PsychBench system includes a dataset, and an evaluation framework built around this dataset. The dataset comprises 300 real cases of psychiatric disorders from three specialized psychiatric medical centers, all documented in Chinese. The evaluation framework addresses one psychiatric domain knowledge evaluation task and five key clinical tasks: clinical text generation, primary diagnosis, differential analysis, medication recommendations, and long-term course management. For each clinical task, specific quantitative evaluation metrics have been developed to ensure scientific rigor and accuracy. During the evaluation, relevant patient information, such as history of present illness and psychiatric examination results, is provided to the LLMs. Detailed instructions, derived from clinical guidelines, are provided to the model, requiring adherence to these prompts when executing the designated clinical tasks. **Fig. 1** presents the primary pipeline of our study. A comprehensive quantitative evaluation of 16 LLMs was conducted using PsychBench. Further assessment examined how factors such as prompt design, chain-of-thought reasoning, input text length, and domain-specific fine-tuning influenced the models' overall performance. Additionally, to identify potential limitations and areas for improvement, a detailed error analysis was performed for each clinical task.

To further explore the practical benefits of existing LLMs as supportive tools for psychiatrists of varying levels of experience, a clinical reader study was conducted. The study included 60 psychiatrists, divided equally into three groups: 20 junior, 20 intermediate, and 20 senior psychiatrists. The experimental design featured two scenarios: one without LLM assistance and one with LLM support. In both scenarios, participants were tasked with completing a set of specific clinical tasks. The study primarily measured the time taken by the clinicians to complete these tasks under each condition, as well as the task performance of each group. Two specialist psychiatrists from a review committee evaluated the participants' task performance based on a predefined scoring criterion.

The establishment of PsychBench offers a scientific foundation for evaluating the practical application of LLMs in psychiatric clinical work. To promote transparency, reproducibility, and ongoing progress in the field, we have made both the dataset and evaluation framework publicly available.

# Results

## Creating the PsychBench dataset and evaluation framework

The construction of the PsychBench dataset and evaluation framework was meticulously designed to ensure scientific rigor and comprehensiveness. For the dataset, a power analysis was conducted to determine the necessary sample size, ensuring statistical significance and reliability of the results. Based on the power analysis and insights from relevant research[12,35], 300 de-identified clinical cases were collected from three geographically diverse and representative medical centers across northern, central, and southern China, ensuring broad regional and cultural representation. These cases incorporated comprehensive information including patients' history of present illness, past treatments, family history, physical and mental status examinations, and ancillary test results, etc. This dataset provided a realistic and detailed clinical context for LLM assessment. An independent expert committee was also established to audit and validate the dataset, ensuring data accuracy and consistency. For detailed procedures on power analysis and dataset construction, please refer to the "Dataset" section in Methods.

In building the evaluation framework, as shown in **Fig. 1**, one psychiatric knowledge evaluation task and five independent clinical tasks were designed based on the dataset: clinical text generation, primary diagnosis, differential analysis, medication recommendation, and long-term course management. These tasks were developed to comprehensively assess the large model's expertise in psychiatric knowledge and its practical application abilities in real-world clinical scenarios. Each task was paired with standard answer given by expert committee, along with specific quantitative evaluation metrics to precisely measure LLM performance across different clinical scenarios. Specifically, each LLM received patient data alongside detailed task instructions, after which it generated outputs aimed at fulfilling the specified objectives. These outputs were rigorously assessed using a combination of general-purpose and psychiatry-specific evaluation metrics, enabling a comprehensive comparison of model performance across various clinical tasks. **Supplementary Tables S1 and S2** provide representative examples for each task, including the prompts, input patient information, reference answers, and sample outputs generated by the evaluated LLMs. For detailed design on each clinical task and associated prompts and evaluation metrics, please see the "Evaluation Framework" and "Quantitative metrics" section in Methods.

## Quantitative evaluation of LLMs using PsychBench

This section presents a detailed report on the quantitative evaluation of 16 mainstream LLMs using PsychBench, focusing on their performance across the psychiatric knowledge evaluation task and five psychiatric clinical tasks: clinical text understanding and generation, principal diagnosis, differential analysis, medication recommendation, and long-term course management. The integrated performance of each model across the six tasks is presented in **Fig. 2-A**. **Fig. 2-B**, meanwhile, summarizes the performance of each model on individual tasks by aggregating multiple evaluation metrics within each task. For more granular rankings and metric-specific comparisons, please refer to **Extended Data Fig. 1–2**, **Extended Data Table 3**, and **Supplementary Table S16**. To gain a deeper understanding of the models' real-world performance, comprehensive error analysis was conducted for each task to identify and explain potential issues in their outputs. Further, the evaluation examined how factors such as prompt design, chain-of-thought reasoning, input text length, and domain-specific fine-tuning influenced the models' outputs. By adjusting these variables, we aim to identify effective strategies for enhancing LLM performance in psychiatric tasks and pinpoint elements that might contribute to errors. These detailed analyses offer valuable insights for future optimization and application of LLMs in psychiatric clinical settings.

**Psychiatric domain knowledge evaluation: LLMs with higher scores on the psychiatric domain knowledge test tended to perform better across the other five clinical tasks, suggesting that adequate domain knowledge is essential for accurate clinical task execution.** In this task, we designed a multiple-choice test incorporating items from the Chinese National Medical Licensing Examination, psychiatric residency program final exams, and authoritative textbooks and guidelines. This task aimed

to capture each model's ability to recall and apply foundational psychiatric knowledge. As illustrated in **Fig. 2-B** and **Extended Fig. 1**, Hunyuan-pro achieved the highest performance on this task with an average accuracy rate of 90.14%, indicating its comprehensive and robust understanding of psychiatric domain knowledge. This advantage on knowledge base translated into the model's consistently strong performance across the broader PsychBench framework: Hunyuan-pro also achieved the highest composite score across all six tasks in the PsychBench (**Fig.2-A**), highlighting the necessity of a solid and broad foundation in domain-specific expertise for LLMs to effectively perform clinical tasks. Concurrently, Qwen-max, which ranked lowest in the knowledge test, demonstrated weak performance in the principal diagnosis, differential analysis, and medication recommendation tasks. This consistent underperformance across knowledge and application tasks further suggests that limited domain knowledge constrains the model's ability to make contextually appropriate clinical decisions, highlighting the intertwined nature of factual knowledge and applied reasoning in psychiatric LLM use cases.

**Clinical Text Understanding and Generation: Current LLMs demonstrate moderate capabilities in understanding and generating psychiatric clinical text, with limited ability to accurately map patient-reported symptom fluctuation characteristics and onset patterns to standardized psychiatric terminology and to extract and articulate key clinical elements with sufficient precision.** This task assessed the capabilities of LLMs in comprehending and generating clinical text. On general summarization metrics, Doubao-pro-32k achieved the highest *ROUGE-L* (44.08±10.73) and *BERTScore* (78.31±4.62), while GPT-4 slightly outperformed in *BLEU* (20.72±11.60). More than half of the evaluated LLMs achieved *BLEU* above 10, *ROUGE-L* above 35, and *BERTScore* above 74. These results suggest that most current LLMs are able to produce clinically coherent summaries. Additionally, most models achieved 100% of *Diagnostic Criteria Completeness Index (DCCI)*, suggesting a robust ability to understand and follow instructions and being capable of outputting the required medical record content modules in full based on prompts. However, the Spark4-Ultra model scored only 95.51±11.28%, which highlights its deficiencies in instruction following and the completeness of generated content.

To evaluate model ability in generating accurate and standardized chief complaints and diagnostic criteria, we defined the indicators *MNER-F1* and *MNER-BERTScore* for quantitative evaluation, which measures the correctness of the generated key information such as the year of the course of diseases, the form of onset, the description of symptom, etc. The Doubao-pro-32k model achieved the highest *MNER-F1*, at 28.36±13.46%, with GPT-3.5-Turbo closely following with an *MNER-F1* of 27.24±15.74%. This indicates that Doubao-pro-32k demonstrates comparatively stronger ability in summarizing and articulating key information such as the course of illness and severity, aligning clinical manifestations with psychiatric terminology more precisely and professionally. However, these relatively low absolute scores on the two metrics indicate that despite partial semantic and content overlap with reference summaries, LLMs often struggle with generating medically accurate and structured descriptions of illness episodes—particularly in mapping nuanced symptom fluctuation characteristics and onset patterns to formal psychiatric terminologies and articulating key clinical elements.

To contextualize these quantitative findings, a structured error analysis across model outputs was performed. As presented in **Extended Data Fig. 6-A** and **Supplementary Table S3**, we classified the errors into four primary categories: course summary errors, onset pattern summary errors, symptom summary errors, and clinical standardization errors. The error analysis revealed that LLMs most frequently struggled with accurately summarizing the onset pattern (40% of total errors) and maintaining clinical standardization (27%), while errors in course summary (25%) and symptom summary (6%) were comparatively less common. Onset errors typically involved misidentifying episodic versus continuous disease trajectories, reflecting difficulties in temporal reasoning. Standardization errors often stemmed from missing required elements or exceeding documentation constraints, suggesting insufficient alignment with psychiatric note-taking conventions. Course summary errors were more frequent in cases with subtle or long-standing symptoms, while symptom summary errors were rare and usually confined to complex or fluctuating presentations. Importantly, as shown in **Extended Data Fig. 7-A and 7-B**, *MNER-F1* and *MNER-BERTScore* were significantly higher in cases without course and symptom summary errors ($p<0.05$), indicating that these metrics effectively capture the precision and clinical appropriateness of key psychiatric elements in model outputs. These findings highlight that while current LLMs show general semantic understanding, they remain limited in accurately mapping patient narratives to psychiatric terminology and in producing structured, clinically compliant documentation.

**Principal Diagnosis: Current LLMs demonstrate limited but varying capabilities in generating**

**accurate principal psychiatric diagnoses, with performance closely tied to their ability to identify and differentiate subtle symptom patterns and follow formal diagnostic criteria.** Notably, GPT-4 achieved the highest Primary Diagnosis Accuracy guided by ICD-10 standards (*ICD10-PDA*) on this task, reaching 56.06±40.41%. This was closely followed by GLM-4 and Gemini-1.5-pro, with an *ICD10-PDA* of 53.14±41.41%  and 53.14±42.24%), respectively. In contrast, Hunyuan-lite lagged considerably with an ICD10-PDA of 31.50±38.07% (detailed in **Extended Data Fig.1** and **Supplementary Table S16**).

These findings indicate that even the best-performing models currently fall short of reliably replicating clinical diagnostic reasoning, particularly in complex psychiatric contexts. Importantly, our evaluation framework deliberately increases the task's complexity by requiring model predictions to match ICD-10 codes the fourth character of the ICD code, such as F31.4 (Bipolar affective disorder, current episode severe depression without psychotic symptoms), and to select from 77 potential diagnostic categories—mirroring real-world psychiatric practice, where accurate subtyping is critical for treatment planning. The *ICD10-PDA* metric reflects a model's capacity to integrate and interpret multimodal clinical information, including symptoms, disease course, and severity, and to map that understanding onto structured diagnostic taxonomies. Therefore, variations in *ICD10-PDA* not only reflect general diagnostic competence but also expose specific deficits in clinical reasoning, symptom differentiation, and terminological precision. The average *ICD10-PDA* across all evaluated models was 48.54%. This unsatisfactory overall performance reflects that accurately identifying psychiatric disorders at the level of diagnostic subtypes remains a challenging task for current LLMs.

To better characterize the nature of diagnostic failures, we categorized errors into four types: symptom assessment errors (64% of total errors), course assessment errors (18%), severity assessment errors (11%), and unclear diagnoses (6%), as shown in **Extended Data Fig. 6-B** and **Supplementary Table S4**. The symptom assessment errors highlight the models' difficulty in recognizing and distinguishing psychotic, depressive, and manic features, particularly when these symptoms present subtly or co-occur. Notably, this task differs from the symptom summarization in the clinical text understanding and generation task by emphasizing the interpretation of current clinical findings rather than the abbreviation and organization of historical symptom descriptions. Further qualitative analysis of error cases suggests two principal causes for these diagnostic inaccuracies. First, complexity of input information, including long, unstructured patient histories or ambiguous symptom timelines, often leads models to overlook or misinterpret salient information. This aligns with our broader findings on the negative impact of input length on performance on this task. Second, the LLMs' lack of detailed mastery and application of psychiatric knowledge, particularly regarding the hierarchical structure and diagnostic criteria of ICD-10, undermines their accuracy in distinguishing among closely related subtypes (e.g., first-episode vs. recurrent depressive disorder).

**Differential Analysis: Current LLMs exhibit substantial limitations in replicating psychiatrists' reasoning in differential diagnosis, particularly in accurately extracting and articulating clinically salient information such as symptom trajectories, disease course, and psychiatric terminology.** While a few models show early promise, the overall performance underscores the urgent need for targeted architectural and fine-tuning strategies to support complex diagnostic decision-making in psychiatry.

To assess the ability of LLMs to perform differential diagnostic analysis, we evaluated their performance using two accuracy-based metrics—$Acc_{main}$ (for the correctness of the principal diagnosis) and $Acc_{diff}$ (for the correctness of the two differential diagnoses). Across all evaluated models, the average $Acc_{main}$ was 48.42%, and the average $Acc_{diff}$ was 29.35%, underscoring the considerable difficulty LLMs face in replicating psychiatric diagnostic reasoning. Among the 16 LLMs evaluated , Doubao-pro-32k achieved the highest $Acc_{main}$ (53.33±48.36%) but did not outperform others in differential diagnosis, ranking fourth in $Acc_{diff}$ (32.58±41.77%). In contrast, Qwen-max demonstrated the highest $Acc_{diff}$ (34.56±40.92%), but its generation quality (*BLEU*, *ROUGE-L*, and *BERTScore*) lagged behind, suggesting that while it is relatively competent at selecting diagnostically plausible alternatives, it struggles with expressing clinical reasoning in structured and semantically coherent form. These discrepancies between structured accuracy and generative quality illustrate a key challenge: high diagnostic precision does not necessarily translate into clinically acceptable analytical reasoning documentation.

Importantly, our findings highlight that successful differential diagnosis requires not only classification correctness but also the accurate articulation of core clinical factors. This is better captured by the *MNER-*

*F1* and *MNER-BERTScore*, which reflect the model's ability to identify and communicate clinically salient named entities (e.g., key symptoms, disease course, psychiatric terms). ERNIE-4-8k achieved the highest *MNER-F1* (28.89±12.93%), and DeepSeek scored highest in *MNER-BERTScore* (89.83±2.41%). As shown in **Extended Data Fig. 7-D**, the *MNER-F1* score for cases correctly answered by the model is significantly higher than the scores for all other error type groups (independent t-test p-value <0.05), with a particularly pronounced difference observed in comparison to the symptom, disease course, and medical history judgment errors groups (independent t-test p-value <0.01). Moreover, as illustrated in **Extended Data Fig. 7-C and 7-E**, the model's *BERTScore* and *MNER-BERTScore* in correct cases are significantly higher than those in error cases due to misinterpreted diagnostic criteria. These significant differences underscore the correlation between precise differential reasoning and correct differential decisions.

Error analysis further supports these quantitative findings. As shown in **Extended Data Fig. 6-C** and **Supplementary Table S5**, the errors in this task can be categorized into several types: symptom judgment errors, disease course judgment errors, misunderstanding of diagnostic criteria, lack of specificity in differential analysis, and omission of medical history information. Similar to previous two clinical tasks, the primary errors are concentrated in the misjudgment of disease course and symptoms, accounting for 33% and 26% of the errors, respectively. Additionally, 21% of the errors stem from the model's misunderstanding of diagnostic criteria. For example, despite clearly identifying recurrent depressive episodes in the patient's history, the model still fails to accurately diagnose recurrent major depressive disorder. Finally, 14% of errors were attributed to the lack of specificity in differential diagnosis, the smallest proportion among all error types. Although this category accounted for the lowest proportion of total errors, its presence underscores the model's limited ability to perform case-specific differential diagnostic reasoning.

**Medication Recommendation: Current LLMs struggle to balance top-choice alignment, precision, and comprehensiveness in psychiatric medication recommendation, with substantial limitations in aligning recommendations with nuanced clinical context, particularly in terms of safety, symptom specificity, and treatment history integration.**

This task evaluates the capability of LLMs in recommending medications within the context of psychiatric clinical practice. Specifically, the task requires the models to provide medication recommendations from the candidate drugs, ranked by recommendation priority from highest to lowest, based on the patient's medical records and various test results. Additionally, the models must articulate the reasons for their recommendations, as well as the co-medication situations if it applies. To capture different dimensions of performance, we used three complementary metrics analogous to common evaluation measures: Top Choice Alignment Score (*TCAS*), similar to accuracy, reflects whether the model's top recommendation matches the actual effective treatment; Medication Match Score (*MMS*), akin to precision, measures the proportion of model-suggested drugs that align with the reference answer; and Recommendation Coverage Rate (*RCR*), analogous to recall, assesses the breadth of clinically appropriate medications included in the model's output. The specific definitions and calculation formulas are detailed in the "Quantitative metrics" section of Methods.

Among the 16 evaluated LLMs, Moonshot-v1-32k achieved the highest average performance. Specifically, it ranked first in *RCR* (43.72±30.99%), second in *TCAS* (13.15±33.79%), and third in *MMS* (35.92±25.48%). As shown in **Extended Data Fig. 1**, apart from Moonshot-v1-32k, no model achieved consistently high rankings across all three metrics. This suggests that current LLMs struggle to balance top-choice alignment (*TCAS*), precision (*MMS*), and comprehensiveness (*RCR*) in psychiatric medication recommendation. Moreover, the absolute performance levels across models remain unsatisfactory. The average *TCAS* across all models was 10.64%, and average *MMS* and *RCR* hovered around 33–36%, highlighting the intrinsic complexity of medication recommendation in psychiatry and the limitations of current LLMs in simulating such decision-making.

In the error analysis of medication recommendations, we assessed the clinical feasibility of the model's suggestions and identified six major categories of errors or inappropriate recommendations: basic medication usage errors, inadequate consideration of adverse drug reactions, errors in combined medication use, overtreatment, lack of reference to the patient's treatment history, and treatment plans conflicting with the current condition. The definitions and examples of each error type were illustrated in **Extended Data Fig. 6-D** and **Supplementary Table S6**. Among these, treatment conflicts with the current condition were the most common, comprising 33% of all errors. This type of error often involved recommending antidepressants to patients with pronounced manic or psychotic symptoms, potentially exacerbating mania or inducing psychotic symptoms. Overtreatment and failure to consider the patient's previous treatment history made up 24% and 14% of errors, respectively, indicating that the model can still improve in accurately incorporating patient history and assessing symptoms. Errors related to basic medication usage, insufficient caution regarding adverse reactions, and combined medication errors accounted for 4%, 18%, and 6% of total errors, respectively. These errors suggest that while the model demonstrates a degree of competency in basic psychiatric pharmacology, evidenced by the relatively low proportion of errors related to fundamental medication usage, there remain important areas requiring refinement to ensure the clinical safety and appropriateness of its recommendations. The evaluation of the medication recommendation task underscored practical challenges in psychiatric pharmacotherapy. The selection of psychiatric medications often requires iterative adjustments to identify the optimal treatment for a patient, resulting in a slower, costlier treatment process. If the model can reliably assist in this area, it has the potential to expedite the medication selection process, reduce clinical costs, and improve the overall treatment experience for patients.

**Long-term Disease Course Management: Current LLMs exhibit decent capabilities in accurately retrieving clinical information and comparing medical variables across timepoints in long-term psychiatric records, but they struggle to understand and summarize symptom fluctuations and evolving disease trajectories.**

This task simulates a common and clinically critical scenario in psychiatric inpatient care: during daily ward rounds, clinicians need to rapidly review multiple prior progress notes, accurately and efficiently search for key information in these records, or analyze the overall evolution trend of the patient's condition. We reformatted real-world clinical records into question-answering (QA) and multiple-choice (MC) formats to simplify the evaluation of the Long-term Course Management task.

Models such as Doubao-pro-32k and GPT-3.5-Turbo outperformed others across both QA and MC tasks, exhibiting relatively strong performance on metrics reflecting semantic understanding and factual alignment. Specifically, the Doubao-pro-32k, which performed best on this task, obtained scores of 32.13±16.49 for *BLEU*, 66.00±16.94 for *ROUGE-L*, 83.20±8.05 for *BERTScore*, 64.83±30.48% for *MNER-F1*, 91.23±7.45% for *MNER-BERTScore* and 88.90±15.87% for *Accuracy*, respectively. It is worth mentioning that the model participating in the evaluation achieved an average accuracy rate of 85.45% on the MC task, and more than half of the LLMs had an accuracy rate exceeding 88%, which demonstrates the model's ability to extract and analyze key information from long-term medical records. This performance indicates a decent proficiency in understanding and processing complex medical data, which is crucial for applications in healthcare where accurate diagnosis and treatment rely heavily on the precise extraction and interpretation of detailed patient histories and long-term clinical interactions.

To gain more fine-grained insights, an error analysis was conducted. Based on question type, we categorized model errors into variation information judgment errors, summary information judgement errors, comparative information judgement errors, locational information judgement errors, and knowledge deficiency caused errors, as shown in **Supplementary Table S7**. The results were illustrated in **Extended Data Fig. 6-E**, which revealed that the majority of model errors stemmed from failures in temporal reasoning and summarization, not factual retrieval or comparison. Specifically, variation information judgment errors (42%) and summary judgment errors (39%) accounted for over 80% of the total errors in this task, indicating difficulty in synthesizing patterns across multiple encounters, such as identifying gradual symptom improvement or relapsing trajectories. By contrast, errors in comparative judgment (3%) and locational recall (5%) were much less frequent, consistent with the models' ability to resolve straightforward comparisons and information retrieval in long text spans. Knowledge-related errors (11%), such as misinterpreting abnormal lab results, reflect that the psychiatric medication knowledge of LLMs still needs to be further enhanced. This discrepancy underscores that current LLMs may appear competent on questions with lower demands for complex reasoning and information integration, yet lack the integrative understanding needed for psychiatric case synthesis. For instance, a large model may correctly retrieve individual symptoms or medication changes, but fail to interpret whether the patient is stabilizing, deteriorating, or experiencing cyclical patterns.

## The influence of different prompt strategies

In our study, we delved into the effects of few-shot and Chain of Thought (CoT) prompting techniques on the performance of LLMs in the five clinical tasks. Specifically, for "Clinical Text Understanding and Generation Task and Differential Analysis" task, we investigated the impact of including or excluding examples in the prompts on model performance. For "Primary Diagnosis Task and Medication Recommendation" task, we examined the influence of employing versus not employing the CoT prompting technique on model performance. The rationale behind this experimental setup is that, in clinical scenarios involving distilling chief complaints, generating structured summaries, and conducting differential diagnostic analysis, clinicians are required to write and organize medical records in a standardized format. In such contexts, the large model must not only ensure the accuracy of concepts and semantics but also adhere rigorously to format standardization. We conducted a few-shot experiments for these two tasks to explore whether the large model can learn the formatting standards for writing mental health clinical records based on a limited number of examples.

**Few-shot prompting facilitates structural format learning but fails to generalize clinical reasoning.** As shown in **Fig. 3-A** to **Fig. 3-D**, our results indicate that providing a single exemplar in the prompts can significantly improve the overall performance of LLMs on "Clinical Text Understanding and Generation" task, particularly those related to formatting standards and language expression habits such as *BLEU*, *ROUGE-L*, and *BERTScore*. This reflects that there is indeed a unique set of formatting standards for psychiatric clinical record-keeping, and the LLMs can effectively master these standards with a small number of examples, thereby aligning its output format with that presented in the examples. Additionally, we observed a corresponding increase in the *MNER-F1* score and *MNER-BERTScore* in the 1-shot setting for the two tasks, indicating that that LLMs can learn the domain-specific formatting and language habits of key elements in psychiatric clinical records with minimal demonstrations.

However, for differential analysis, the use of 1-shot prompts did not bring improvement and even caused a slight decrease in the metrics measuring the correctness of the final differential diagnosis decision. Specifically, as shown in **Fig.3-D**, in the 0-shot and 1-shot settings, the average $Acc_{main}$ of obtained by the evaluation model on the "Differential Analysis" task is basically unchanged. Moreover, it is noteworthy that the accuracy rate of differential diagnosis ($Acc_{diff}$) even slightly decreased in the 1-shot scenario, which may be due to the exemplar inducing a bias in the model's differential diagnostic choices. This discrepancy reveals that while LLMs can be nudged toward stylistic conformity, they struggle to emulate the reasoning patterns necessary for accurate differential diagnosis. The slight decline in $Acc_{diff}$ further suggests that exposure to exemplars may inadvertently induce diagnostic bias, highlighting a key challenge when deploying few-shot methods for reasoning-heavy psychiatric clinical tasks.

**Chain-of-thought (CoT) prompting did not improve the performance on psychiatric reasoning tasks that require professional complex inference and longitudinal decision-making.** For "Primary Diagnosis" task and "Medication Recommendations" task, in real clinical practice, clinicians engage in multi-step complex and implicit logical reasoning when making diagnoses and developing treatment plans. These reasoning processes are not detailed in a fixed format within the text of medical records. Consequently, we designed CoT comparative experiments for these two tasks to investigate whether the LLM can enhance the quality of diagnostic and treatment suggestions by simulating the thought processes of clinicians.

The results of the CoT prompting comparative experiments are depicted in **Fig. 3-E** to **Fig. 3-G**. For primary diagnosis, the use of CoT-style prompts led to a decrease in the *ICD10-PDA* of primary diagnosis. Similarly, for medication recommendation, the average performance of LLMs on the metrics of *TCAS*, *RCR*, and *MMS* decreased after using the CoT-form prompt. The average *TCAS* dropped from 10.65% to 9.15%, *RCR* from 38.11% to 33.62%, and *MMS* from 35.27% to 28.53%. Upon manual inspection of the responses provided by the model under the CoT-form prompt, we found that CoT outputs of LLMs often included more detailed reflections on medical records, such as symptom evolution and prior medication outcomes. However, these insights did not translate into accurate diagnosis or appropriate therapeutic adjustments. When suggesting medication, LLMs tended to recommend drugs that had appeared in the medical records, despite their previous suboptimal treatment outcomes, showcase in **Supplementary Table S13**. In contrast, the recommended medications by psychiatric clinicians in the standard answers showed a lower overlap with previously used drugs, with a preference for adjusting medications to achieve better therapeutic effects. Additionally, we observed that even when LLMs identified a drug's poor efficacy in the analysis phase, they failed to make correct and reasonable adjustments in the final medication recommendation, which highlights the current LLMs' insufficient reasoning ability in transitioning from past medication and efficacy analysis to drug adjustment plans. These findings indicate that in the field of psychiatric practice, while CoT prompting encourages the model to simulate reasoning steps in appearance, it does not instill true clinical reasoning capability.

Psychiatric decision-making often requires integrating nuanced symptom trajectories, treatment responses, and diagnostic uncertainties. These are capacities that current LLMs cannot reliably simulate even with the help of reasoning prompts, underscoring the need for specialized training.

## The influence of input length

Compared to common nature language processing (NLP) tasks, psychiatric clinical tasks often involve lengthy and complex input texts[36,37]. As illustrated in **Fig. 4-A**, the five tasks in *PsychBench* exhibit broad input length distributions, necessitating LLMs to possess fine-grained extraction and analysis capabilities and manage long-range dependencies. **Fig. 4-B** illustrates the mean values of various evaluation metrics for LLM groups with different context window lengths across 5 clinical tasks. Based on the model's context length, the LLMs tested were categorized into four groups: $8k$, $32k$, $128k$, and $> 128k$. Since only GPT-3.5-turbo has a context length of $16k$, it was not included in this analysis. In the following, we conduct a detailed analysis (**Fig. 4-C**) of how model performance varies with input length across different tasks, highlighting that increased context window size does not uniformly translate into better performance, particularly for psychiatric clinical tasks requiring complex reasoning and integrative clinical understanding.

It is evident that on "Clinical Text Understanding and Generation" task, as input length increases, the performance of models with various context lengths shows a trend of first slightly decreasing, then rising and decreasing again. This suggests that the relationship between input length and performance may not be linear, and certain models may perform better with specific input lengths. This could be attributed to the fact that longer inputs might introduce additional complexity or noise, which affects the model's ability to generate concise and accurate outputs. On the other hand, shorter inputs may not provide enough context for the model to generate a comprehensive and accurate chief complaints or diagnosis criteria. The task demands both extraction of relevant details and the ability to generate a coherent response within the constraints of clinical standards, making it essential for the model to balance brevity with completeness.

On "Primary Diagnosis" task, as the input length increases, LLMs with a context length of $8k$ exhibit a consistent decline in performance, whereas models with a context length of $32k$ or more show a trend of decreasing performance followed by an increase. This phenomenon reflects the trade-off between the difficulty of information extraction and analysis in long texts and the richer information provided by more detailed patient information and medical records for diagnosis. For LLMs with shorter context lengths, the increase in input length results in a richer set of diagnostic information, but the model lacks the capacity to extract it effectively; for models with longer context lengths, when the input exceeds 6000 words, the positive impact of the additional information outweighs the negative effects of analyzing longer texts, leading to an upturn in overall performance. It can also be observed from the polyline graph of this task that LLMs with longer context show a more significant performance improvement after the inflection point.

For “Differential Analysis” task, as input length increases, the performance of all models showed an upward trend regardless of context length. For LLMs with a context length of $32k$ or more, the performance bound is more significant after the point of 4500-5000 input length, with the 128k LLM group with the longest context window achieving the best average performance on this task for input lengths >4500.

For “Medication Recommendation” task, contrary to the “Differential Analysis” task, the performance of all four groups of context length models showed a decreasing trend as the input length increases. This downward trend suggests that LLMs struggle to fully analyze and understand more detailed and complex historical medication and disease progression records. Despite longer inputs providing richer information, the participating LLMs are unable to effectively utilize this information to assist in reasoning and make better medication recommendations. At the same time, in clinical practice, developing the next step in the treatment plan for patients with chronic, recurrent conditions and extensive historical medication records is indeed a more challenging task. Additionally, it should be noted that the input length distribution of “Differential Analysis” task and “Medication Recommendation” task is different, the input length of the latter is mostly distributed between 2500 and 3500, while the input length of the former is mostly distributed between 3500 and 4500, which may also explain the opposite relationship between the performance of the model and the input length on these two tasks.

In “Long-term Course Management” task, which includes both QA and multiple-choice subtasks, two distinct mechanisms emerged. For the QA subtask, as input length increases, the performance of LLMs with context length of 32k or more trends upwards, with models having a context length greater than $128k$ showing a more rapid performance improvement before the input length reaches 4000 words, indicating some benefit from contextual breadth. In contrast, the group of models with an 8k context window exhibit a performance trend that first decreases and then increases, with the turning point also occurring in the 3000-4000 words range. In the MC subtask, the performance of all groups of models with different context lengths generally shows a downward trend as the input length increases. As shown in **Extended Data Fig. 8**, for models with a context window greater than or equal to 32k, the accuracy on the multiple-choice questions of this task fluctuates with the position of the correct answer in the medical records, showing a trend of first decreasing and then increasing. Notably, when the answer is located at a relative position of 0.2-0.4 in the medical records, the accuracy decreases most significantly. This phenomenon is closely related to the "lost in the middle"[38] effect, suggesting that models with longer context windows tend to lose focus on key information in the middle part of the text when processing long documents, leading to a noticeable decline in accuracy in the middle section.

In conclusion, these results underscore that the current strategies for extending LLM context length may impair their analytical and reasoning abilities, as the 5 clinical tasks designed by PsychBench require not only the extraction of key information but also a certain level of understanding and analysis of the input content combined with psychiatric expertise. Other studies have also found that after extending the context window, LLMs do not necessarily “understand” the content better[39], and model performance is influenced by the position of the answer within the input[40]. These results alert us to reconsider the current strategies for extending LLM context length and the methods of evaluation.

## The comparison between general-purpose LLMs and LLMs fine-tuned on medical domain

To enhance the capabilities and adaptability of LLMs in the medical field, numerous efforts have been made to fine-tune general-purpose LLMs using medical literature, medical encyclopedias, or consultation records from internet hospitals, thereby constructing medical-specific LLMs. For instance, HuaTuoGPT2 is a medical large model fine-tuned based on the general-purpose model Baichuan2. In this evaluation, we conducted a performance comparison between HuaTuoGPT2 and Baichuan2 under both 0-shot and 1-shot scenarios. The relative capabilities of the two models across five tasks are illustrated in **Fig. 4-D**. In the heatmap presented, colors are determined based on the comparative ratio of HuaTuoGPT2 to Baichuan2 on specific performance metrics, with red hues indicate that HuaTuoGPT2 outperforms Baichuan2 in terms of the specific metric, while blue hues suggest that Baihcuan2 has the advantage. The depth of coloration corresponds to the magnitude of the performance differential.

It is evident that for the five tasks designed for PsychBench, the fine-tuned HuaTuoGPT2 in the medical domain demonstrates a nuanced superior or comparable performance on most metrics compared to the general-purpose model Baichuan2. This advantage is more pronounced in terms of the $Acc_{main}$ in differential diagnosis tasks and the *MMS* of medication recommendations in supportive treatment decision-making tasks. These results indicate that fine-tuning in the medical domain can bring about a subtle improvement in the overall performance of LLMs in psychiatric clinical diagnosis and treatment tasks. Moreover, the experimental results also reveal that in both 0-shot and 1-shot scenarios, the fine-tuned HuaTuoGPT2 in the medical domain exhibits slightly inferior performance than the general model Baichuan2 or shows no advantage over it in terms of the *ICD10-PDA* of principal diagnosis and the *TCAS* and *RCR* rate of medication recommendations, compared to the general-purpose model Baichuan2. **Fig. 4-E** presents a comparison of the performance evaluated by *BLEU*, *ROUGE-L*, and *BERTScore* of the two models on tasks 1, 3, and 5, which involve the composition of summaries and analytical texts. The figure is structured such that the vertical axis denotes the scores achieved by Baichuan2 for the respective metrics, while the horizontal axis represents the corresponding scores for HuaTuoGPT2. Each data point within the plot corresponds to an individual test case from the benchmark. The distribution of points within the upper half of the quadrant would signify that Baichuan2 attains superior scores to HuaTuoGPT2 across a greater number of test cases, and conversely, a concentration in the lower half would imply a superior showing by HuaTuoGPT2. The results reveal that the data points are almost evenly distributed on both sides of the dash line, which represents equivalent performance between the two models. This observation is further supported by the heatmap, where the colors corresponding to these metrics are relatively light, trending towards white, suggesting a lack of clear distinction in performance between HuaTuoGPT2 and the general-purpose model Baichuan2. In other words, despite the targeted fine-tuning of HuaTuoGPT2 in the medical domain, it does not demonstrate a clear advantage over the general-purpose large model in the psychiatric clinical tasks designed by PsychBench.

## Reader study

Psychiatric clinical work heavily relies on clinical experience, leading to differences in performance among psychiatrists with varying levels of experience when completing clinical tasks. Therefore, to more thoroughly examine the effectiveness of LLMs in assisting psychiatrists at different experience levels, and to further analyze the potential strengths and limitations of LLMs to provide directions for future research, we designed and conducted a clinical reader study. We recruited 60 psychiatrists with varying levels of experience: 20 junior, 20 intermediate, and 20 senior psychiatrists. **Extended Data Fig. 3** illustrates the detailed design of the reader study. Participants were asked to complete a series of clinical tasks (including diagnosis, differential analysis, and medication recommendations) under two conditions: with and without LLM assistance. Subsequently, specialist psychiatrists evaluated their responses to compare the performance between the two scenarios, as well as across different experience levels. The scoring criteria specifically for the reader study were developed based on ICD-10 guidelines, as shown in **Extended Data Table 4**. The reader study user interface is presented in **Extended Data Fig. 4**.

As depicted in **Fig. 5**, the assistance of existing LLM had varying effects on psychiatrists with different levels of experience. A substantial improvement was observed in the overall performance of junior psychiatrists, with average overall scores increasing from 22.85 to 26.25 (p-value = 0.013). Psychiatrists with intermediate and higher levels of seniority demonstrated slight performance enhancement, with average overall scores rising from 26.35 to 27.9 (p-value = 0.276) and from 29.0 to 30.2 (p-value = 0.242), respectively.

In the diagnostic task, the results of the reader study indicated that physician groups with different levels of experience performed well in completing the task, consistently providing correct diagnoses (scoring 5 points), as shown in **Fig. 5-B**. However, analysis of the violin plot shapes revealed that the lower half of the LLM-assisted group was narrower compared to the group without LLM assistance. This change suggests that the assistance of LLMs has, to some extent, reduced the likelihood of incorrect diagnoses in the diagnostic task. Notably, the effect of LLM assistance was more pronounced in the lower- and mid-experience groups. This finding indicates that for less experienced psychiatrists, LLMs can provide valuable support and reference, helping to reduce errors and biases during the diagnostic process.

The goal of differential diagnosis is to analyze and differentiate potential similar diseases based on the patient's clinical condition. Differential accuracy mainly measures the hit rate of identifying important potential diseases after considering the specific clinical context. As shown in **Fig. 5-B**, the accuracy of differential diagnosis was relatively lower in the low-experience group. This is primarily because less experienced psychiatrists often lack specificity when performing differential analysis. For example, in the case of depressed patients with delusions or anxiety, psychiatrists should consider differential diagnoses such as delusional disorder and generalized anxiety disorder. Psychiatrists in the low-experience group sometimes overlooked these possible similar conditions. With the assistance of LLM, the lower bound of the differential accuracy in the junior group improved, although it did not reach a statistically significant difference (p-value = 0.16). The differential completeness mainly measures the ability to conduct a thorough analysis of potential diseases. The results in **Fig. 5-B** indicate that LLM assistance significantly improved the comprehensiveness of differential diagnosis in junior and intermediate groups. The effect of LLM assistance is primarily reflected in its ability to provide a detailed analysis based on the patient's clinical condition. Psychiatrists can quickly reference this content to capture the patient's condition more effectively and develop a clear differential thought process, thereby delivering accurate and comprehensive differential analyses and reducing the risk of missing potential diseases.

In psychiatric practice, there is often no single correct treatment plan, with multiple clinically appropriate pharmacological options potentially available for the same patient. Therefore, we evaluated medication recommendations from physicians with varying levels of experience from multiple dimensions. First, in terms of medication accuracy, physicians in the intermediate and senior groups performed significantly better than those in the junior group. The deficiencies in the junior group primarily stemmed from insufficient analysis of the patient's symptoms progression and treatment history. For example, as presented in **Supplementary Table S9**, a junior psychiatrist failed to recognize that a patient with treatment-resistant depression and anxiety had been on an adequate dose of venlafaxine for six months with poor efficacy, and did not offer any medication options to address the patient's anxiety symptoms.

While benefiting from LLM's detailed analysis of the patient's condition and treatment history, along with its provided medication suggestions, the average score of medication accuracy of junior group increased by 14%. However, in terms of medication adherence to clinical guideline, the effect of LLM assistance was negligible across all experience groups. This result reflects the limitations of current LLMs in adhering to clinical medication guidelines. The primary issue is the model's lack of reliable, real-world clinical guideline knowledge and practical experience, which impedes its performance in strictly following specific treatment protocols. Thus, while LLMs can provide valuable medication recommendations, there remains room for improvement in ensuring that these recommendations fully comply with clinical standards and treatment guidelines. In the evaluation of medication contraindication accuracy and comprehensiveness, junior group showed improvement with the assistance of LLM, with accuracy and comprehensiveness increasing by 14% (p-value = 0.18) and 19% (p-value = 0.08), respectively. The primary contribution of LLMs was providing detailed interpretations of the patient's condition and relevant test results. Furthermore, LLMs assisted physicians by offering knowledge about drug interactions and contraindications, helping to reduce the risk of prescribing medications that are contraindicated. This is especially valuable for junior psychiatrists, as they may lack sufficient experience when managing complex cases. However, for more experienced physicians, the effect of LLM assistance was negligible. Nevertheless, LLM still contributed by offering a rapid analysis of the patient's condition, which can enhance efficiency in clinical decision-making. Therefore, we performed a statistical analysis of the efficiency across the different groups.

In terms of productivity, as presented in **Fig. 5-C**, the LLM has been shown to markedly reduce the time psychiatrists require to formulate primary diagnoses, conduct differential diagnoses, and devise medication regimens. For the junior group, the average time to process a case was 535.7 seconds, which was significantly reduced to 292.6 seconds with LLM assistance, indicating the most pronounced efficiency gains. The intermediate group demonstrated the highest efficiency, with an average case processing time of 337.0 seconds, further reduced to 217.4 seconds with the aid of LLM. Conversely, the senior group exhibited a less pronounced reduction in average case processing time, decreasing from 524.2 seconds to 399.6 seconds with LLM assistance. Notably, the efficiency gains attributed to LLM for the senior group were not statistically significant, with a p-value of 0.705, suggesting that the impact of current LLMs on the workflow of senior psychiatrists may be limited.

## Discussion

This study conducted an in-depth analysis of the application of LLMs in the field of psychiatric clinical practice. As the prevalence of mental disorders rises, the traditional psychiatric clinical practice faces increasingly evident challenges. The emergence of LLMs presents new possibilities for addressing these critical issues. However, the actual effectiveness of LLMs in psychiatric clinical practice has yet to be thoroughly validated, which limits their practical application and hinders further research on LLMs tailored for psychiatric applications. To address this gap, we have developed a benchmarking framework—PsychBench—grounded in real clinical data, standardized clinical guidelines, and the actual demands of clinical practice. The PsychBench is designed to comprehensively evaluate the performance of LLMs in psychiatric clinical settings, providing robust evidence for the reliable assessment of their efficacy in real-world applications, and guide future research in this area.

The PsychBench framework stands out for several reasons. Firstly, unlike previous evaluations that only focused on single tasks such as conversational coaching[26,27], and therapy behavior simulation[25], its design acknowledges the distinctiveness of psychiatric practice by decoupling and defining clear, clinically significant sub-tasks with customized evaluation indicators. This approach ensures that the evaluation of LLMs is aligned with the practical demands of psychiatric clinical practice, something that general medical benchmarks have failed to achieve. Secondly, rather than using non-clinical social media blogs[28] or generating simulation data through simulation[24] or rewriting[25], the framework is grounded in high-quality, annotated data from real-world clinical scenarios. This ensures that the evaluation indicators can effectively and objectively measure LLM performance, capturing the subtleties of mental health assessments that are often missed by previous benchmarks. Thirdly, PsychBench offers a practical and easy way to comprehensively evaluate the capacity of LLMs in psychiatric clinical practice, which respond to the urgent needs of the current research community[29]. By defining task-specific prompts and quantitative evaluation metrics for each clinical task, PsychBench enables multidimensional assessments that are both efficient and thorough. This structured approach facilitates a nuanced understanding of model performance across various aspects of psychiatric care. Utilizing PsychBench, we evaluated the psychiatric clinical performance of 16 LLMs varying with respect to open-source properties, manufacturers, number of parameters, and specific domains, obtaining an advanced and holistic view of LLM's strengths and challenges in the field of psychiatric clinical practice.

LLMs have demonstrated advantages in clinical text comprehension and generation tasks, particularly in structured summarization. Leveraging few-shot learning, LLMs can quickly adapt to new format requirements with minimal example support. This learning paradigm reduces training costs while enhancing task execution flexibility. For instance, by providing one single example, most LLMs can rapidly learn to generate structured summaries adhering to the formatting conventions and narrative styles commonly used in psychiatric medical records. This capability of LLMs provides a potential approach to assist psychiatrists in saving time on tedious documentation while ensuring that the generated text meets clinical needs. However, we observed that LLMs may occasionally exhibit errors in summarizing and mapping complex clinical information to standardized psychiatric terminology, such as inaccurate characterization of disease progression or misinterpretation of key symptoms. Therefore, further optimization of LLMs is necessary to ensure they meet the higher accuracy standards required for clinical documentation applications. Promising directions include efficient fine-tuning strategies such as QLoRA and advanced in-context learning approaches, which have shown potential in general medical applications[12] but remain underexplored in psychiatry.

LLMs exhibit inadequate performance in diagnostic tasks and currently do not meet the clinical demands for accurate diagnoses. In psychiatry, such demands include high accuracy, consistent reasoning, and transparent justification. However, current LLMs fall short of these standards: the top model achieved only 56.06% accuracy in principal diagnosis and 32.58% accuracy on identifying two most plausible differential diagnoses—substantially below the reference level of 77.5% established by intermediate psychiatrists group in our reader study. The underlying reason is that most models evaluated are general-purpose, lacking in-depth training specific to psychiatric clinical expertise. As a result, they often demonstrate insufficient mastery of domain-specific knowledge and underdeveloped clinical reasoning abilities. Furthermore, the requirement for the models to provide definitive diagnoses from among 77 possible subtypes of mental disorders undoubtedly complicates the diagnostic process. Although this requirement poses challenges for the models, it more accurately reflects the complexities and diversity inherent in psychiatric diagnosis in real clinical settings. The error analysis revealed that most diagnostic errors made by LLMs are concentrated on the misinterpretation of patient symptoms. This limitation stems from the complexity and variability of symptoms in psychiatric patients, where accurate diagnosis requires not only precise identification of the patient's current symptoms, but also thorough analysis and comprehensive consideration of the patient's past symptom and diagnosis changes. This highlights the importance of fine-tuning LLMs with clinical data, as current models primarily possess guideline-based knowledge but lack extensive clinical experience. Targeted learning on authentic and complex clinical data may help bridge this gap, improving diagnostic accuracy and enabling more reliable performance in complex, real-world scenarios[41], thereby supporting the emergence of new paradigms in psychiatric diagnosis and clinical decision-making[33].

LLMs can provide psychiatrists with helpful differential analyses assistance. Although existing models still exhibit limitations in their decision-making abilities for accurate diagnoses, their robust text comprehension and analytical capabilities enable them to generate detailed differential analyses based on patient information and specific instructions. The reader study indicated that LLMs offer particularly significant support to junior psychiatrists, helping them access more thorough differential analysis references and thereby improving diagnostic accuracy. This auxiliary function not only enhances their work efficiency but also bolsters their capacity to handle complex cases, making new paradigms for diagnosis and assessment possible[33]. Error analysis showed that the primary issues still stem from inaccuracies in assessing patients' medical history and symptomatology.

For medication recommendation, while current LLMs face challenges in optimizing top-choice alignment, precision, and comprehensiveness in psychiatric medication recommendations, they nonetheless hold promise as supportive tools to enhance clinical decision-making. The LLM can first conduct a thorough analysis of the patient's condition and relevant auxiliary test results, and then, by integrating knowledge of relevant medications and diseases, generate an appropriate medication recommendation. For instance, LLMs can suggest appropriate combinations of antidepressants and antipsychotics to a patient with slight psychotic symptoms, while some junior psychiatrists tend to focus solely on the patient's depressive symptoms, neglecting potential psychotic symptoms. In the reader study, the LLM demonstrated strong supportive effects, providing substantial references for doctors in formulating medication plans. The errors in medication recommendations made by the LLM predominantly fall into three categories: treatment plans conflicting with current condition, overtreatment, and inadequate consideration of adverse drug reactions. These error patterns underscore the potential clinical risks of deploying LLMs as autonomous tools for psychiatric medication management, highlighting the need for fine-tuning the model with real clinical data to enhance its adherence to clinical guidelines and experiential knowledge. By incorporating further specialized knowledge training and employing techniques such as retrieval-augmented generation (RAG) [42,43], the practicality and safety of LLMs in individualized medication recommendations can be enhanced.

LLMs possess the capability to rapidly retrieve target information from lengthy texts, a feature that holds significant value in the long-term management of patients with mental disorders. This ability enables models to swiftly integrate patients' historical medical records, treatment histories, and symptom changes, thus aiding psychiatrists in extracting key information from complex datasets to optimize clinical decision-making processes. The LLM demonstrated a high accuracy rate in this task. However, this capability has certain limitations. First, it depends on the model's context window size; if a patient's medical history exceeds the model's processing capacity, critical information may be overlooked or inadequately utilized. Given the "lost in the middle" phenomenon, simply increasing the context window length does not effectively address this challenge. Additionally, error analysis reveals that current LLMs still struggle to understand and summarize symptom fluctuations and evolving disease trajectories, ultimately affecting the effectiveness of long-term patient management—a concern also highlighted in a study on the use of large language models in transforming chronic disease management[44].

We explored multiple factors that may influence LLM performance to provide guidance for subsequent work. First, in-context-learning (ICL) can significantly improve model's ability to adhere to domain-specific formatting conventions and expression styles, even if only one example is used for model adaptation. However, the impact of more examples on performance was not investigated due to the long context of the clinical tasks involved in this study exceeds the limits of some models' context windows. Moreover, we investigated the effect of input context length on model performance. A key observation is that the relationship between input length and model performance is not linear, and the optimal model context window length appears to vary by task and input length. The results highlight the challenges LLMs face when dealing with long and complex clinical notes, requiring fine-grained extraction and inference capabilities that are often affected by the context window of the model. This complex interplay between input length and LLM performance across psychiatric clinical tasks reveals that longer context window length or inputs do not necessarily lead to improved outcomes. While extended context windows allow models to process more comprehensive information, they also introduce challenges related to information extraction and coherence, particularly in tasks that require fine-grained analysis and reasoning.

The CoT-based prompting strategy did not necessarily improve model performance on specific tasks. Our experiments suggest that the use of CoT-style prompts even led to a decrease in the performance of primary diagnosis and treatment planning. A similar situation was found in a study by Yang et al. on ChatGPT's ability to perform mental health analysis and emotional reasoning tasks[45]. The reason is that in the clinical domain of psychiatry, the physician's thought-decision process, which is constructed over a long period of extensive clinical practice, is multifaceted, nonlinear, and somewhat personalized. It is infeasible to construct chains of thought to boost model's performance by simply adjusting input prompt to improve model performance for the specialized domains and complex tasks. Conversely, its low-quality or erroneous analyses may lead to greater biases. A more feasible way to inject clinical decision-making reasoning into a model is to fine-tune the model using real clinical data. However, it is important to note that fine-tuning needs to be targeted. As shown in **Fig. 4-D** and **Fig. 4-E**, HuatuoGPT2, which has been fine-tuned with medical data, does not perform significantly better than the generic model Baichuan2 on PsychBench. This discrepancy may arise due to the unique terminology norms and diagnostic decision-making processes in psychiatric clinical practice. These knowledge and logic cannot be acquired through low-quality, broad medical internet corpora. Instead, it necessitates the collection and organization of high-quality clinical corpora, including real-world clinical case records, authoritative guidelines, and cutting-edge academic papers, etc.

In order to further assess the effectiveness of existing LLMs as auxiliary tools for psychiatrists of different experience levels, we conducted a clinical reader study. The results revealed that, in terms of work quality, LLMs did not significantly improve the performance of senior and intermediate psychiatrists. However, they notably enhanced the clinical performance of junior psychiatrists, particularly in the comprehensiveness of differential diagnosis analysis and medication recommendations. Regarding work efficiency, LLMs demonstrated significant improvements for both junior and mid-level psychiatrists, boosting their ability to complete clinical tasks more efficiently. These findings highlight the potential value of LLMs in supporting psychiatric clinical work and underscore the differing needs of psychiatrists at various stages of their careers. This differences suggest that future LLM development should consider tailoring LLM assistance based on the doctor's level of expertise and area of specialization, to more comprehensively support the development of psychiatric practice. For example, for junior psychiatrists, the LLMs should focus on supporting foundational knowledge and disease management, while for more experienced psychiatrists, the assistance can be more centered on the latest research findings and updates to clinical guidelines.

We also identified the following limitations of our study. First, we only investigated the performance of the original models and those adapted using in-context learning (ICL). We did not explore more advanced fine-tuning methods such as LoRA, primarily due to the substantial volume of high-quality labeled data required to fine-tune large models for psychiatric-specific clinical tasks. While recent efforts have produced LLMs for mental health support using synthetic or internet-sourced data (e.g., for emotional support or online consultation) [46,47], there remains no publicly available or clinically validated LLM fine-tuned on real-world clinical data for psychiatric diagnostic and treatment practice. Nonetheless, our findings demonstrate that existing generic LLMs already show promise in assisting psychiatrists with core clinical tasks. In future work, we aim to build on this foundation to develop domain-adapted LLMs that are fine-tuned specifically for psychiatric clinical practice.

Second, while the dataset was constructed from three authoritative psychiatric centers across China and covers diverse regional and ethnic populations, it is entirely in Chinese, which may limit its immediate applicability in non-Chinese settings. That said, all cases were diagnosed using ICD-10, a globally accepted standard, and the dataset is available in both Chinese (original form) and English (upon request). Third, the dataset exhibits a long-tailed ICD distribution, with some diagnoses underrepresented. This reflects real-world psychiatric prevalence rather than artificial balancing, and enhances ecological validity. However, it may challenge model performance on rare disorders. In future iterations of PsychBench, we plan to expand coverage of rare disorders and edge cases to improve the robustness and fairness of evaluation. Together, these limitations suggest important directions for future work, including international collaboration, cultural adaptation, and the development of fine-tuned psychiatric models.

In conclusion, this study proposes a benchmark for evaluating the performance of LLMs in assisting psychiatric clinical practice, known as PsychBench, which includes a dataset and an evaluation framework. Through quantitative assessments of existing LLMs and a clinical reader study, we identify the potential of these models to assist psychiatric clinician. However, despite demonstrating certain potential advantages, LLMs exhibit significant shortcomings and do not fully meet clinical application needs. These deficiencies primarily manifest in diagnostic accuracy, application of specialized knowledge, and handling of complex cases—particularly those requiring nuanced reasoning across longitudinal symptom trajectories, comorbidities, and individualized patient presentations. Overall, this research provides a systematic evaluation framework and reference for the future development of LLMs in the psychiatric field, underscoring the importance of further optimizing LLMs to achieve greater clinical adaptability and effectiveness.

# Methods

## Dataset

To ensure the validity and reliability of our research findings, we first determined the required sample size through power analysis. In this study, we established the following parameters: Effect Size: Based on literature and previous studies[48,49], we assumed an effect size of 0.5, defined using Cohen's d, which is considered a medium effect. Alpha Level $(\alpha)$: We selected 0.05 as the significance level to control the risk of Type I errors. Statistical Power $(1-\beta)$: We set this to 0.90, indicating our aim to have a 90% chance of detecting a true effect, thereby reducing the risk of Type II errors. We utilized the 'statsmodels' library in Python to perform the power analysis calculations, which yielded a required sample size of 85.03. This means we need to collect data from at least 86 patients for the assessment in this study. A power analysis curve is presented in **Extended Data Fig. 5**. This systematic approach to sample size calculation ensures that our research possesses adequate statistical power, thereby enhancing the credibility and generalizability of the results.

During the data collection phase, we collaborated with three prestigious psychiatric medical centers to ensure diversity and representativeness of the benchmark dataset, including Beijing Anding Hospital affiliated with Capital Medical University, Fourth People's Hospital of Wuhu, and Second People's Hospital of Dali. Based on the sample size calculations, we determined that at least 86 patient data points were necessary. Ultimately, we collected clinical patient data from each center, totaling 300 cases, with 100 cases from each facility. This approach not only exceeds the requirements established by our power analysis but also provides a broader context and richer data characteristics for the study.

Beijing Anding Hospital, a National Medical Center for Mental Disorders located in northern China, is a high-complexity, Grade-III-A tertiary psychiatric hospital with over 800 beds and more than 860,000 annual outpatient and emergency visits, representing the most severe and complex psychiatric cases. Fourth People's Hospital of Wuhu, in central China, serves both as a major referral center and a primary psychiatric care provider for the regional population, with 1,760 beds and over 150,000 outpatient visits annually, reflecting both specialized and provincial-level psychiatric services. Second People's Hospital of Dali, a Grade-III-A tertiary hospital situated in a multi-ethnic autonomous prefecture of southwest China, covers a catchment area of over 3 million people and provides culturally informed mental health care, particularly to ethnic minority populations. The inclusion of these institutions ensures broad coverage across clinical complexity levels, referral pathways, and cultural contexts.

The selection of these three medical centers covers different geographical regions and medical backgrounds, providing a solid foundation for the representativeness of the data. In addition, patients of multiple ethnic minorities were collected, whose living habits, cultural backgrounds, and health beliefs vary significantly. These differences may not only have a potential impact on the pathogenesis and clinical presentation of psychiatric disorders but also influence communication methods and medication adherence during the treatment process. By including data from ethnic minority patients, this study offers a more comprehensive representation of psychiatric patient characteristics across different cultural backgrounds. This diversity of culture and habits adds richer dimensions to the construction of evaluation benchmarks, ensuring the applicability of the research findings across various ethnic and cultural contexts. Importantly, all diagnostic labels in the dataset are based on the ICD-10 system, which is globally widely used and provides a standardized framework for psychiatric diagnosis. This adherence to psychiatric practice standards further enhances the generalizability of the dataset beyond the Chinese clinical context. The study was reviewed and approved by the ethics committees of all participating institutions according to the data verification and annotation guidelines (**Extended Data Fig. 9**), ensuring compliance with local cultural norms and ethical standards.

We implemented several key screening standards to ensure data quality and the validity of the study. First, we focused on inpatients admitted after 2022. This criterion was designed to exclude clinical data biases introduced by the COVID-19. Second, to reflect the actual clinical scenarios of different medical centers, we selected case data by referencing the statistical distribution of patient diagnoses at each center, focusing primarily on schizophrenia and mood disorder spectrum conditions. This selection process ensured not only the authenticity of the dataset but also the diversity of disease types, providing a robust foundation for evaluating the performance of LLMs in psychiatric clinical tasks. The patient data we collected encompasses essential information, including demographic details (such as age, gender, and occupation), history of present illness, past medical history, personal history, family history, and treatment history. Additionally, complete clinical records from the hospitalization process were included, which contain multiple physical examinations, psychiatric assessments, results of auxiliary examinations, medical orders, and physician ward round notes. To protect patient privacy, all data underwent strict de-identification procedures prior to analysis, including the removal of direct identifiers (e.g., names, addresses, contact information) and manual inspection of free-text fields to eliminate any residual sensitive information. Only retrospective clinical data were used, and all data were fully anonymized before researchers accessed them. The study protocol received approval from the Ethics Committee of Beijing Anding Hospital, Capital Medical University. The individual data were completely anonymous, making the study exempt from requiring informed consent. An independent expert committee in psychiatry reviewed and validated the collected data to ensure its accuracy and reliability. Detailed statistics of the dataset are presented in **Extended Data Table. 1**.

## Evaluation framework

The construction of the evaluation framework comprises three components: the design of evaluation tasks, the formulation of quantitative metrics for each task, and the design of prompts for each task. It includes one task specifically aimed at assessing psychiatric domain knowledge, alongside five clinically grounded tasks: clinical text understanding and generation, principal diagnosis, differential analysis, medication recommendations, and long-term disease course management. In the following sections, we describe each task in detail, providing examples of the prompts used and elaborating on the intended evaluation goals.

**Task 0: Psychiatric domain knowledge evaluation.** Assessing the foundational and applied psychiatric knowledge of LLMs is critical for ensuring their safe and effective integration into clinical workflows. Unlike general medical knowledge, psychiatric expertise requires nuanced understanding of abstract concepts such as symptom phenomenology, diagnostic thresholds, treatment appropriateness, and ethical considerations—areas where model hallucinations or insufficient mastery of domain knowledge may pose significant clinical risks. To this end, we developed a domain-specific knowledge test designed to rigorously evaluate the model's comprehension and application of psychiatry-relevant knowledge. Specifically, we curated a domain-specific knowledge test dataset consisting of 639 multiple-choice questions (MCQs) drawn from three complementary sources. The first subset comprises 131 MCQs selected from the psychiatry section of the CMExam[50], a benchmark dataset based on the Chinese National Medical Licensing Examination. These questions reflect standardized assessments of core psychiatric knowledge required for clinical certification. The second subset includes 273 MCQs obtained from final examinations used in psychiatric residency training programs across multiple institutions in China. These items emphasize real-world clinical reasoning and decision-making, thereby assessing the model's ability to apply theoretical knowledge to complex patient care scenarios. The third subset consists of 235 MCQs curated from authoritative psychiatric sources including the DSM-5, ICD-10, peer-reviewed clinical guidelines, and widely used psychiatric textbooks. These questions evaluate the model's understanding of diagnostic criteria, classification systems, and evidence-based treatment standards.

For this task, in the prompt, we require the LLM to act as a professional psychiatrist and complete the MCQs, each of which contains a single correct answer. The model is required to select and output only the correct option without providing any explanatory reasoning or analysis. **Supplementary Table S1** presents the prompt used in this task and an example of input question.

**Task1: Clinical text understanding and generation.** This task requires the model to extract and generate chief complaints and structured summaries that adhere to clinical standards from detailed patient information. The input consists of comprehensive patient information, including demographic details (age, gender), medical history (present, past, personal, family, and treatment history), and results from physical, mental, and auxiliary examinations after hospitalization. The model is tasked with generating a first course record with two main parts: (1) a concise chief complaint, summarizing the patient's symptoms and clinical course in no more than 20 words, and (2) a structured summary covering four specific dimensions: symptoms, clinical course, severity, and exclusion criteria. Unlike conventional text summarization, the chief complaint must be concise yet informative, capturing the essence of the patient's condition. The structured summary must accurately reflect the patient's symptoms, the progression of the disease, its severity, and any exclusions (i.e., conditions that have been ruled out based on the patient's history and diagnostic data).

For this task, the prompt delineated the standards for writing clinical chief complaints, adhering to clinical medical record conventions. We required the LLM to limit the chief complaint to no more than 20 words. For the structured summary, the prompt states detailed explanation of the four aspects: symptoms, course of disease, severity, and exclusion. **Supplementary Table S1** presents the prompt used in this task and an example of input patient information.

**Task2: Principal diagnosis.** This task focuses on fine-grained psychiatric diagnosis, requiring the model to provide precise diagnoses based on detailed patient information. The input patient information is identical to that used in Task 1. The model is required to adhere strictly to the ICD-10 diagnostic criteria and provide a diagnosis refined to the fourth character of the ICD code, such as F31.4 (Bipolar affective disorder, current episode severe depression without psychotic symptoms). This requirement necessitates that the model not only classify the primary psychiatric disorders but also conduct more nuanced subtype diagnoses based on the patient's specific condition. This diagnostic process is more complex than standard diagnostic tasks and closely reflects actual clinical scenarios, where the specific type of disease is significant for the subsequent development of treatment plans. This task can assess model's understanding and application of clinical diagnostic standards, as well as its ability to navigate complex patient scenarios that require nuanced diagnostic decisions.

For this task, the prompt specified that the primary diagnosis should be based on the ICD-10 diagnostic criteria and required the model to refine the diagnosis to the disease subtype. The ICD-10 diagnostic codebook was embedded in the prompt, thereby guiding and constraining the outputs of LLMs to align with standardized medical nomenclature. Specifically, we listed 77 common psychiatric disorders along with their corresponding ICD-10 codes in the prompt, thereby standardizing and limiting the model's diagnostic outputs. **Supplementary Table S1** presents the prompt used in this task and an example of input patient information.

**Task3: Differential Analysis.** This task is designed to evaluate the capability of LLMs in conducting differential diagnosis within the domain of psychiatry, a process that is both cognitively demanding and clinically indispensable. Given the high degree of symptom overlap across psychiatric disorders, patients presenting with similar clinical features may in fact suffer from distinct conditions that require different treatment strategies. Accurate differential diagnosis is therefore critical to avoiding misdiagnosis and ensuring appropriate care. For instance, distinguishing between bipolar affective disorder, which involves alternating manic and depressive episodes, and major depressive disorder, characterized solely by depressive episodes, is essential for informing therapeutic decisions and preventing inappropriate interventions. In this task, the LLM is instructed to generate one primary diagnosis that best reflects the patient's actual condition, alongside two differential diagnoses that represent clinically plausible alternatives. The differential diagnoses should be grounded in a comparative analysis of the patient's symptoms, signs, medical history, and examination findings, thereby simulating the reasoning process of a trained psychiatrist. This design choice acknowledges the diagnostic ambiguity that frequently exists in early psychiatric assessments, where a precise diagnosis may not be immediately evident. Instead, clinicians often consider a spectrum of possible conditions that require further evaluation through longitudinal observation, additional history-taking, or targeted investigations. By mirroring this diagnostic uncertainty and prompting the model to explore multiple plausible hypotheses, the task is intended to assess the model's ability to reason under uncertainty, navigate broad diagnostic categories, and support clinicians in complex diagnostic scenarios.

For this task, we designed the prompt to instruct the model to analyze comprehensive patient information and provide one primary diagnosis along with two differential diagnoses. The prompt explicitly specified the expected diagnostic output format and guided the model to organize its analysis consistent with standard psychiatric differential diagnostic procedures. To standardize and constrain the model's primary and differential diagnostic choices, we supplied a list of 26 common psychiatric disorders along with their ICD-10 codes, covering a broad range from F00 to F98 in ICD-10. Importantly, in this task, the ICD-10 categories were presented at a coarser granularity, typically up to the integer before the decimal point (e.g., F32 for depressive episode), and in some cases, as code ranges (e.g., F70-F79 for intellectual disabilities), to reflect the variable specificity often encountered in psychiatric differential diagnosis practice. **Supplementary Table S1** presents the prompt used in this task and an example of input patient information.

**Task4: Medication recommendation.** This task necessitates LLMs to prescribe the optimal psychiatric therapeutic medication based on the medical context and disease progression of the patients. To simulate this decision-making process, LLMs are required to provide optimal therapeutic medications alongside recommendation reasons, grounded in the analysis of multiple clinical inputs, including the patient's present illness history, past medical history, personal and family psychiatric history, prior treatment records, as well as findings from initial physical examination, mental status assessment, and relevant auxiliary tests conducted post-admission. Medication recommendations should be presented with clear prioritization, and rationales must reflect an understanding of both clinical indications and potential contraindications. For instance, when evaluating a patient with comorbid arrhythmia and major depressive disorder, the model is expected to avoid recommending tricyclic antidepressants such as amitriptyline, due to their known risk of exacerbating cardiac conduction disturbances. This task is crucial for assessing the model's ability to integrate complex clinical data, apply pharmacological knowledge, and make safe, individualized treatment decisions, thereby offering insight into its real-world utility in augmenting psychiatric clinical care. Given the non-uniqueness of correct answers in medication recommendation, ground truth labels were expert-annotated to include multiple clinically appropriate options based on real-world prescriptions, ensuring a comprehensive and fair evaluation.

For this task, the prompt requires the model to analyze the provided comprehensive patient information and offer medication recommendations in order of priority. Based on clinical practice guidelines, the prompt outlined the factors and strategies that the model should consider during medication recommendation, with a particular emphasis on the careful evaluation of drug interactions and adverse reactions. To reduce task complexity and standardize the output of medication names, a predefined list of 34 commonly prescribed psychiatric medications was provided, thereby constraining the candidate drug space and ensuring that recommendations remained psychiatric relevant. The list of alternative disease names and drug names provided in the prompts of Task2-4 contains all the diseases and drugs covered in the standard answer. **Supplementary Table S1** presents the prompt used in this task and an example of input patient information.

**Task5: Long-term course management.** This task is designed to evaluate an LLM's ability to understand, retrieve, and reason over temporally extended and clinically rich psychiatric records, simulating core aspects of long-term, multi-stage clinical interactions. Unlike traditional static assessments that rely on isolated medical knowledge question answering or decontextualized patient information in a single time section, this task presents models with full-course hospitalization records that unfold over time, structured chronologically by date and encompassing daily clinical notes, physical examination findings, psychiatric evaluations, and auxiliary test results. Each patient case spans multiple time points and reflects a dynamic evolution of psychiatric symptoms, physical examination findings, and auxiliary test results, mirroring the longitudinal and iterative nature of real-world psychiatric care. To operationalize this complexity in an evaluable form while preserving clinical realism, we constructed both open-ended question answering (QA) and multiple-choice (MC) tasks. Each question targets temporally anchored and context-dependent aspects of the patient's course, such as "How did the patient's psychiatric symptoms present after the initial MECT session on Day 10?" or "In the recent auxiliary examination, which indicator was higher than the reference value but was not mentioned in the previous examination? A. Glutamate aminotransferase B. Aspartate aminotransferase C. Triglycerides D. High-density lipoprotein". For each patient profile, we extracted three tailored questions from their extensive longitudinal medical records, and the ground-truth answers were directly extracted from original records during evaluation data preparation. Through this task, the model's ability to retrieve and analyze long-term hospitalization information in real time can be evaluated, which enables clinicians to quickly identify information like key medication responses and changes in the patient's condition during hospitalization, allowing for more timely and context-aware adjustments to treatment plans.

For this task, the prompt instructed the model to thoroughly examine the patient's multi-stage course records, inspection results, and other information. We then require the model to provide answers to these questions in one session. This design simulates the real-world clinical scenario in which psychiatrists must efficiently review a patient's evolving medical history during each ward round and rapidly extract key information to inform treatment decisions. **Supplementary Table S1** presents the prompt used in this and an example of input patient information.

## Quantitative metrics

We have designed detailed evaluation criteria for the five tasks within the PsychBench. The guiding principle for the design of these metrics is to enable quantitative assessment of the performance of specific tasks based on their characteristics, and these metrics can be calculated automatically.

*BLEU*, *ROUGE-L*, and *BERTScore* are traditional metrics more commonly used in machine translation and summarization tasks. However, they are limited in their capacity for the medical domain. These scores reflect the degree of structural and lexical similarity between the generated text and the provided reference, but they do not specifically assess critical information such as symptom descriptions, medication usage, disease names, anatomical and physiological terms, and laboratory tests in diagnostic and treatment contexts. The challenge is particularly significant in the psychiatric medical domain, where generating diagnosis and treatment plans often involves navigating abstract concepts, precisely grasping and defining symptoms, paying particular attention to past medication and complications, and dealing with omissions and hallucinations (fabrication, falsification, and plagiarism). In response to this issue, PsychBench further proposes evaluation metrics based on medical named entity recognition: the *Medical NER F1 Score (MNER-F1)* and the *Medical NER BERTScore (MNER-BERTScore)*, which assess the quality of key information in LLM outputs from the perspectives of strict keyword matching and keyword semantic similarity, respectively. Specifically, in line with the approach outlined by Bureaux Tao et al. (https://github.com/Bureaux-Tao/ccksyidu4k-ner), we developed a specialized medical named entity recognition (NER) model, termed $M_{NER}$, optimized for the analysis of medical electronic health record notes. This model was trained on the CHIP2020 dataset, which encompasses 2.2 million characters, 47,194 sentences, and 938 documents, with an average document length of 2,355 characters. The dataset includes a diverse range of medical entities across nine major categories, such as 504 common diseases, 7,085 anatomical names, 12,907 clinical manifestations, and 4,354 medical procedures. To enhance the model's applicability to the psychiatric and psychological domain, we assembled a separate collection of psychiatric clinical electronic medical records, distinct from the 300 cases in PsychBench. We meticulously annotated a training set of psychiatric and psychological electronic medical record notes, with a focus on domain-specific entities and expressions. This annotation process involved the identification and labeling of entities such as mental disorders, psychiatric treatments, and psychological terminology. Subsequent to this annotation, we fine-tuned the NER model on this specialized dataset. The resulting $M_{NER}$ model, based on this dataset, demonstrates an F1-score of 0.66 for the identification of medical entity keywords, indicating a robust performance in recognizing relevant entities in the psychiatric and psychological domain. If we denote the reference answer for the $i$th case in the benchmark as $r_i$ and the LLM's output as $o_i$, we have:

$$M_{NER}(r_i) = \{en_{r_i 1}, en_{r_i 2}, \dots, en_{r_i m}\}$$

$$M_{NER}(o_i) = \{en_{o_i 1}, en_{o_i 2}, \dots, en_{o_i n}\}$$

$$Medical\ \ NER\ \ Precision = \frac{|M_{NER}(r_i) \cap M_{NER}(o_i)|}{|M_{NER}(o_i)|}$$

$$Medical\ \ NER\ \ Recall = \frac{|M_{NER}(r_i) \cap M_{NER}(o_i)|}{|M_{NER}(r_i)|}$$

$$Medical\ \ NER\ \ F1 = 2 \times \frac{Medical\ \ NER\ \ Precision\ \ \times\ \ Medical\ \ NER\ \ Recall}{Medical\ \ NER\ \ Precision +\ \ Medical\ \ NER\ \ Recall}$$

Where $\{en_{r_i 1}, en_{r_i 2}, \dots, en_{r_i m}\}$ represents the m medical entities predicted after inputting $r_i$ into $M_{NER}$. || denotes the number of elements in the set. Considering that the reference answers and LLM outputs may have slightly different descriptions for the same symptoms, we use MedicalBERT to further compare the semantic similarity between the two sets of identified named entities at the semantic level. Specifically, we have:

$$MedicalBERT\left(en_{r_i j}\right) = v_{r_i j} \in R^{1\times d}$$

$$MedicalBERT\left(M_{NER}(r_i)\right) = V_{r_i} = \ \left[v_{r_i 1}, v_{r_i 2}, \dots, v_{r_i m}\right] \in R^{m\times d}$$

$$MedicalBERT\left(M_{NER}(o_i)\right) = V_{o_i} = \ \left[v_{o_i 1}, v_{o2}, \dots, v_{o_i m}\right] \in R^{n\times d}$$

$$Medical\ \ NER\ \ BERTScore = \frac{1}{m}\Sigma_{k=1}^{m} \max_{j}\left(cosine\ \ similarity\left(V_{r_i}, V_{o_i}\right)\right)$$

In task1 (Clinical text Understanding and Generation), task3 (Differential Analysis), and task5 (Management of Long-Term Disease Progression), which involve summary, information extraction and analysis, we use these two metrics to provide a more comprehensive quantitative evaluation of model performance.

Below, we will introduce the evaluation metrics for each of the six tasks in detail:

**Psychiatric domain knowledge evaluation.** In this task, LLMs are required to act as psychiatrists to complete MCQs. The questions are derived from the Chinese National Medical Licensing Examination, final examinations used in psychiatric residency training programs, and authoritative psychiatric sources. These MCQs have only one correct option for each question. We use the average *Accuracy* of the model in completing all the questions as the evaluation metric of this task.

**Clinical text Understanding and Generation.** This task encompasses two key components: the abstraction of the patient's chief complaint and the synthesis of structured summary through comprehensive analysis of the patient's information and medical history. The chief complaint serves as a concise abstraction of the patient's narrative, while deriving the structured summary necessitates a thorough analysis and synthesis of the patient's medical data. Therefore, in this benchmark, we use the commonly employed *BLUE*[51], *ROUGE-L*[52], and *BERTScore* metrics for the distillation of the chief complaint and the generation of the structured summary. Additionally, the description of the disease courses in the chief complaint and the analysis and summary of the structured summary involves the calculation of time and the mapping of symptoms to psychiatric professional descriptions. Hence, in addition to evaluating the summarization ability through *BLUE*, *ROUGE-L*, and *BERTScore*, this benchmark also includes *Diagnostic Criteria Completeness Index (DCCI)*, *MNER-F1* and *MNER-BERTScore* to assess the integrity and accuracy of the generated structured summary, respectively. Specifically, the percentage of answers generated by the LLM that cover all four diagnostic criteria — "symptom criteria", "disease course criteria", "severity criteria", and "exclusion criteria"—is used as the *Diagnostic Criteria Completeness Index (DCCI)* metric. In addition to the aforementioned metrics, this task also employs the *MNER-F1* and the *MNER-BERTScore* as evaluation indicators. These metrics are specifically designed to assess the quality of medical named entity recognition in the outputs of LLMs.

**Principal Diagnosis.** This task aims to evaluate the LLM's ability to process complex patient information and provide a primary diagnosis. The model's output includes the International Classification of Diseases, 10th Edition (ICD-10) [53] codes and their corresponding disease names. To comprehensively measure the diagnostic accuracy of the model, we use *ICD-10 guided Primary Diagnosis Accuracy (ICD10-PDA)* as the evaluation metric, calculated based on the overlap between the model's predictions and the reference ICD-10 codes. The method is as follows: if the LLM's predicted ICD-10 code exactly matches the reference answer, it indicates that the model successfully predicted the disease category and sub-type, and the case's *Accuracy* is scored as 1. If the first three digits of the LLM's predicted ICD-10 code match the reference answer but differ from the fourth digit onwards, it shows that the model correctly predicted the disease category but failed to precisely identify the sub-type, scoring 0.5 for that case. If the first three digits of the LLM's predicted ICD-10 code do not match the reference answer, the model is considered to have failed in diagnosing the disease category, and the *Accuracy* is scored as 0. This fine-grained accuracy calculation method allows us to evaluate the model's performance more comprehensively in disease diagnosis tasks, reflecting its strengths and weaknesses in recognizing different disease categories and sub-types.

**Differential Analysis.** The objective of this task is to evaluate the LLM's capability in distinguishing between potential psychiatric diagnoses. The model is tasked with accurately identifying the primary diagnosis and suggesting two most probable differential diagnoses, supported by a comprehensive analysis and rationale derived from the patient's information. Additionally, the model should highlight the key differential aspects of the primary and differential diagnoses under consideration. To quantify the LLM's efficacy in pinpointing the primary diagnosis from the choices presented in the prompt, we employ the $Acc_{main}$ metric. For each case, if the LLM's identified primary diagnosis aligns with the reference diagnosis, the case is assigned an $Acc_{main}$ score of 1; otherwise, it receives a score of 0. To evaluate whether the model's proposed differentials align with expert-provided reference answers, which represent the most clinically relevant alternative possibilities for each case, we utilize the $Acc_{diff}$ metric. For each case, the model receives 0.5 points for each correct differential diagnosis (up to a maximum of 1.0). To gauge the depth and quality of the LLM's differential diagnosis analysis, we compute the *BLEU*, *ROUGE-L*, and *BERTScore* metrics; to specifically measure the LLM's grasp of key information and the accuracy of its analysis of key symptoms, we calculated the *MNER-F1* and *MNER-BERTScore*. These metrics are used to compare the LLM's analytical output with the reference analysis provided by clinical experts in the fields of psychiatry and psychology. This comparative analysis ensures a comprehensive assessment of the LLM's diagnostic reasoning capabilities.

**Medication recommendation.** This task requires the LLM to provide medication recommendations based on the patient's medical history and various test results, in order of recommended priority, along with an explanation of the reasons. To rigorously assess the concordance between the LLM's medication suggestions and the reference answer, we have developed a set of evaluation metrics grounded in the hit rate concept. Let $D_{LLM} = \{d_{l1}, d_{l2}, \dots, d_{ln}\}$ be the set of medications recommended by the LLM and $D_{ref} = \{d_{r1}, d_{r2}, \dots, d_{rn}\}$ be the set of medications recommended by the reference answer. The medications recommended in both sets are sorted in order of recommendation priority from high to low. The metrics to evaluate LLMs performance on this task are defined as follows:

*Recommendation Coverage Rate (RCR)*: This metric is calculated as the ratio of the number of medications recommended by the LLM that are also present in the benchmark response's recommended medications list.

$$Recommendation\ \ Coverage\ \ Rate = \frac{|D_{LLM} \cap D_{ref}|}{|D_{ref}|}$$

Recommendation Coverage Rate quantifies the exhaustiveness of the LLM's recommendations.

*Medication Match Score (MMS)*: Medication Match Score is determined by dividing the number of medications from the benchmark response that are correctly identified by the LLM by the total number of medications suggested by the LLM. The formula for Medication Match Score is:

$$Medication\ \ Match\ \ Score = \frac{|D_{LLM} \cap D_{ref}|}{|D_{LLM}|}$$

Medication Match Score gauges the exactness or appropriateness of the LLM's medication suggestions.

*Top Choice Alignment Score (TCAS)*: Accuracy is a measure of whether the LLM's highest-ranked medication corresponds with the top medication in the benchmark response. The formula for Top Choice Alignment Score is:

$$Top\ Choice\ Alignment\ Score = \begin{cases} 1 & if\ \ d_{l1} = d_{r1} \\ 0 & if\ \ d_{l1} \neq d_{r1} \end{cases}$$

This metric evaluates whether the LLM's top-ranked medication aligns with the top choice in the reference answer, showcasing the reliability of the LLM in critical decision-making.

**Long-Term course management.** This task necessitates that the LLM accurately extract and comprehend information from patients' longitudinal medical record texts to complete custom-designed reading comprehension and multiple-choice questions.

In the reading comprehension subtask, three sets of reading comprehension questions and answers were generated for each medical record, with a focus on details such as changes in the patient's condition, adjustments to treatment plans, and examination results. This task leans more towards comprehending summaries, where the model needs to read and analyze the long texts of medical records produced during a patient's prolonged hospitalization. The goal is to accurately capture the specific information queried in the questions, interpret the information within the medical records using its own psychological and psychiatric domain knowledge, consider clinical examination indicators, and analyze the medication and its effects. The evaluation metrics for this task include *BLEU*, *ROUGE-L*, and *BERTScore*, to measure the accuracy and fluency of the LLM's responses against the reference answers, as well as *MNER-F1* and *MNER-BERTScore* to measure the precision in identifying key entities in the responses. These two sets of evaluation metrics together provide standardized and precise quantitative assessment.

In the multiple-choice subtask, 4 multiple-choice questions were generated for each medical record, with each question having only one correct option among the four to five options provided. These questions target specific information such as medication dosages on particular days within the long-term case records and specific numerical values of certain indicators from an examination, thereby examining the LLM's ability to accurately extract specific information from lengthy texts. The evaluation metric for this task is the average *Accuracy* rate of the LLM's responses to the multiple-choice questions.

For ease of interpretation, **Supplementary Table S14** provides a simplified overview of all evaluation metrics used in PsychBench, including their definitions, computational approaches, and clinical significance.

## LLMs

To ensure a comprehensive and representative evaluation of LLMs in psychiatric clinical tasks, we selected a broad set of popular models that vary in open-source properties, manufacturers, scale, training background, and intended application. This diverse set of LLMs enables us to investigate the general capabilities, strengths, and limitations of LLMs in supporting psychiatric practice. Considering the extensive textual inputs involved, we selected models with a minimum context length of 4*k* tokens or more, ensuring they can process lengthy clinical texts effectively. We organized the evaluated LLMs into two categories:

**General-Purpose Models.** This group includes both Chinese-developed and internationally developed models that are not specifically fine-tuned for the medical domain.

- **Chinese general-purpose LLMs:** We selected state-of-the-art models developed by leading Chinese AI companies, including ERNIE4-8k[54], Hunyuan-pro[55], Huanyuan-lite[55], Doubao-pro-32k[56], GLM4[57], Qwen2.5 (Qwen-max) [58], and Spark4-Ultra[59], as well as other widely used domestic large models, including Deepseek-chat-v2[60], Moonshot-v1-32k[61], Baichuan4[62], Yi-large[63], and MiniMax[64]. These models are typically trained with a focus on Chinese-language optimization, making them especially relevant for our dataset and clinical context.
- **Multilingual models with global prominence:** We also included internationally recognized multilingual LLMs such as the GPT series (GPT-3.5-turbo[65], GPT-4o-mini[66], GPT-4[67]) and Gemini-1.5-pro[68]. These models are among the most widely used models worldwide and have demonstrated strong cross-linguistic generalization abilities. Including these models allowed us to perform cross-comparisons and assess the extent to which globally leading models can generalize to complex psychiatric clinical tasks.

This broad inclusion of general-purpose models allows us to analyze performance across language backgrounds, modeling strategies, and deployment scenarios. By comparing a wide range of models, we aim to identify consistent patterns in performance as well as task-specific challenges that persist across model families.

**Medical Domain Fine-tuned Large Models.** To compare the capabilities of large models fine-tuned in the medical domain with those of general-purpose large models in assisting psychiatric diagnosis and treatment, this evaluation also includes state-of-the-art medical domain fine-tuned large models. To conduct a fairer comparison, we selected Baichuan2-7B-Chat[69], which was fine-tuned on general data, and HuaTuoGPT2[70], which was fine-tuned on medical data, both based on the Baichuan2-7B-base model. This pairing enabled a controlled analysis of whether medical-domain fine-tuning improves task performance in psychiatric contexts.

The names, parameter sizes, and context lengths of the large models involved in the evaluation are shown in **Extended Data Table. 2**. We conduct experiments of Baichuan2-7b-base and HuaTuoGPT2 on a single NVIDIA Tesla A100 GPU with 80GB of memory. The results of all the rest of the LLM experiments are obtained by calling the corresponding API. Each prompt is fed independently to avoid the effects of dialogue history.

## Prompt strategy

Based on the prompts specifically designed for each clinical task, we employed three proven prompt strategies to guide the model in completing psychiatric clinical tasks and compared their final performance across different tasks.

**Zero-shot learning** does not rely on any examples but instead directly depends on the task description and contextual information for reasoning[71]. In this study, we used the answers generated by the model using the zero-shot learning strategy as a baseline to assess its ability to handle tasks without any prior examples. Additionally, zero-shot learning was also used as a reference standard for evaluating the effectiveness of other prompt strategies.

**Few-shot learning** helps the model better understand the context and task requirements by providing a small number of examples, without adjusting the model's weights, thereby improving its performance on specific tasks[71]. In this study, we applied few-shot learning to Task 1-4 (Clinical Text Understanding and Generation, Principal Diagnosis, Differential Analysis and Medication Recommendation), and detailed analyzed the performance of the LLMs on Task1 (Clinical Text Understanding and Generation) and Task3 (Differential Analysis) using 0-shot and 1-shot prompting strategies. These tasks have high requirements for the content and format of the model's output. By providing a small number of examples, the model could more precisely understand the specific requirements of the task and effectively capture the relationship between input and output, thereby enhancing task completion and accuracy. It is important to note that due to the length of the patient information, including multiple examples in the prompt could exceed the context window limit of some models. Therefore, in this study, we only used one example for few-shot learning to ensure the model's context window limit was not exceeded. Regarding the selection of examples, research indicates that choosing relevant examples can effectively enhance model performance[72]. However, to ensure fairness in evaluation, we selected random samples as examples for testing.

**Chain of Thought** (CoT) strategy is an approach designed to guide the model through step-by-step reasoning, helping it draw more logical conclusions when facing complex tasks[71]. In this study, we applied the CoT strategy to Task 2 (Primary Diagnosis) and Task 4 (Medication Recommendation). These tasks require the model to perform diagnostic reasoning and medication suggestions based on the clinical information provided. This not only demands strong information processing capabilities but also requires detailed and rigorous reasoning. By explicitly guiding the model through a step-by-step reasoning process, the CoT strategy theoretically enhances the model's accuracy and rationality when handling complex information, enabling it to make better diagnostic and medication recommendations. However, although the CoT strategy can provide some level of guidance in the form of instructions within the prompt, the actual effectiveness still largely depends on the model's inherent reasoning ability. Therefore, in some complex clinical tasks, the application of the CoT strategy may not significantly improve the model's performance, particularly when handling intricate clinical decision-making scenarios. In Primary Diagnosis task, we constructed the reasoning chain based on the ICD-10 diagnostic criteria, guiding the model to first analyze the patient according to the ICD-10 standards and then provide a diagnosis. In Medication Recommendation task, we developed a reasoning chain based on several factors, including the patient's condition, symptoms, diagnostic and examination results, adverse drug reactions, drug interactions, and adherence to treatment protocols. The model was required to perform a thorough analysis of these aspects before providing a final medication recommendation. Examples of the prompts used to implement the CoT strategy in both tasks are provided in the **Supplementary Table S15**.

## Experiments on Input Length

To investigate how input length affects model performance across different psychiatric clinical tasks, we conducted a dedicated analysis using the full set of test samples from PsychBench. For each case across the five clinical tasks, the total input length, calculated as the combined number of tokens in the fixed task-specific prompt and the variable patient-specific information, was measured. Based on the input token counts, we grouped the cases into input length bins to analyze performance trends. For each group, we computed the mean scores of the task-specific evaluation metrics. This allowed us to assess how model performance varies as a function of input length within each task.

To better understand the scalability of different LLMs, we categorized the models into four groups according to their maximum context window sizes: 8k, 32k, 128k, and >128k. Since only GPT-3.5-turbo has a maximum context length of 16k, it was excluded from this specific analysis for consistency. Through this setup, we aimed to capture how different models handle varying input lengths and whether extended context capabilities translate to improved performance in real-world psychiatric tasks involving long and complex clinical inputs.

## Reader study

After quantitative evaluating the LLMs performance in completing psychiatric clinical tasks using automated metrics, we designed and conducted a reader study to thoroughly assess the application of LLMs as assistive tools for doctors with varying levels of experience, thereby providing more insight for further development of related research. The specific design of the reader study is illustrated in **Extended Data Fig. 3**. We primarily analyzed the study results from two perspectives: work quality and efficiency.

Unlike purely quantitative evaluations, the reader study offers a more realistic reflection of the practical utility and potential of LLMs in clinical practice.

During the preparation phase of the study, we began by recruiting participants. A total of 60 psychiatric psychiatrists were recruited for the study, including 20 psychiatrists each from three experience levels: junior (less than 5 years of experience), intermediate (5-10 years of experience), and senior (more than 10 years of experience). Additionally, we invited two specialist psychiatrists from an independent expert committee to serve as evaluators. One expert was responsible for scoring the participants' responses, while the other conducted a review of the scores to minimize potential biases and maintain reliability of assessment. This experimental design not only allows us to analyze the auxiliary effects of LLMs across psychiatrists with varying levels of experience but also helps us understand the potential development directions for LLMs in real-world clinical applications.

The development of scoring criteria was a critical component of the reader study. To ensure scientific rigor, objectivity, and reproducibility, we designed detailed scoring standards based on the ICD-10 diagnostic guidelines, with reference to frameworks SaferDx[30]. These standards cover multiple dimensions, including (1) Diagnostic Accuracy: The correctness of the diagnosis provided; (2) Differential Accuracy: The precision in differentiating between similar conditions; (3) Differential Completeness: The thoroughness of the differential diagnosis process; (4) Medication Accuracy: The correctness of prescribed medications; (5) Medication Adherence to Guidelines: Adherence to standard protocols in medication prescription; (6) Contraindication Accuracy: The avoidance of contraindicated medications in the prescription; (7) Contraindication Completeness: The comprehensiveness of identifying and avoiding contraindications. Each dimension was accompanied by clear scoring guidelines and corresponding point definitions, as illustrated in **Extended Data Table 4**. The scoring standards underwent review and revision by an independent expert committee to ensure comprehensiveness and consistency. This thorough validation process guarantees that the criteria can be used reliably in future studies, facilitating comparison and reproducibility across different research efforts. Additionally, we will create a public leaderboard to showcase the performance of different models to encourage further research and advancements in this area.

The specific execution process of the reader study is detailed in **Extended Data Fig. 3**. A subset of 100 patient cases was randomly selected from the entire dataset for reader study. Psychiatrists were divided into three groups based on their experience levels (junior, intermediate, and senior) and completed tasks under two conditions (without LLM assistance and with LLM assistance), resulting in a total of six groups, each comprising 20 psychiatrists. Each psychiatrist was required to complete 10 cases out of the 100, ensuring that each case was repeated twice in each group and appeared a total of 12 times across all groups. For each case, psychiatrists needed to accomplish three primary tasks: primary diagnosis, differential analysis, and medication recommendation. We excluded tasks related to text summarization and case management, as these primarily involved text processing and did not effectively assess psychiatric expertise. In the no-LLM assistance condition, psychiatrists provided answers independently; in the LLM assistance condition, they modified their answers by referencing the responses generated by the LLM before finalizing their responses. After collecting responses from the six groups, the expert psychiatrists would score the responses, and the scoring results would be aggregated by group. Additionally, psychiatrists were required to record the time spent on each task to analyze the impact of LLM assistance on work efficiency. This design ensures the rigor and the reproducibility of the study.

## Error analysis

To better understand model limitations and support fine-grained evaluation, we conducted a systematic error analysis for each clinical task in PsychBench. This analysis serves to complement quantitative evaluation metrics and provide clinically meaningful insights into model performance.

For each task, we selected the top-performing LLM based on composite scores to undergo manual error annotation. Errors were categorized based on the nature of the task, typical clinical reasoning failures, and empirical patterns observed in model outputs. Two independent annotators with expertise in psychiatry and computational linguistics conducted the labeling. Discrepancies were resolved through consensus discussion with senior psychiatrists to ensure clinical validity. Detailed definitions and representative examples of each error category are provided in **Supplementary Tables S3–S7**, corresponding to the five clinical tasks, respectively.

## Statistics analysis

In determining the sample size, we referred to similar studies and incorporated statistical power analysis. We collected 100 clinical cases from each of the three medical centers, with a uniform distribution of samples from each center, totaling 300 cases for the evaluation dataset. During the power analysis, we followed relevant guidelines and literature recommendations to set the effect size and statistical power parameters, avoiding reliance on determining the effect size through pilot studies, as this approach is considered unreliable and may waste data[73]. We believe the sample size is sufficient as it enabled reproducible and highly credible results when conducting the same experiment with a different set of samples.

For all the selected LLMs, we adopt their default hyper-parameters to maintain consistency with standard operational settings. To ensure the generation of deterministic responses, the temperature parameter was configured to 0.1. Additionally, to prevent premature termination of responses, the maximum token limit for new generations, denoted as $max_new_tokens$, was set to 4096, thereby ensuring the integrity of the generated text.

We applied min-max normalization to revalue each evaluation metric for every task and then calculated the mean of all metrics for each task as the overall performance indicator. Subsequently, we computed the mean of the overall performance indicators across the six tasks to serve as the comprehensive evaluation metric for the large models in the field of psychiatric care. In the Result section, the specific scores of LLMs on each indicator are presented in the form of "mean ± standard deviation".

In the quantitative evaluation, we found that GPT-4 exhibited the highest diagnostic accuracy in the diagnostic tasks. This highlights GPT-4's ability to accurately analyze patient conditions and its deep understanding and application of psychiatric clinical knowledge. Accurate diagnosis is the cornerstone of psychiatric clinical practice and serves as the foundation for developing subsequent treatment plans. Considering its overall performance across all tasks, we selected GPT-4 as the LLM-assisted tool for the reader study to comprehensively evaluate its effectiveness in supporting psychiatrists with varying levels of experience in real-world clinical tasks.

In the reader study, we randomly selected 100 cases for comparative analysis. The study was conducted by 60 psychiatrists with varying levels of experience, who were evenly distributed into three groups based on their experience levels: junior, intermediate, and senior psychiatrists, with 20 psychiatrists in each group. To ensure the fairness and scientific integrity of the experiment and adhere to the principle of repetition, each of the 100 cases was completed and evaluated by 6 psychiatrists (2 psychiatrists from each of the 3 experience-based groups). Specifically, each psychiatrist was assigned 10 cases, following a structured allocation: cases 1–10 were assigned to the first psychiatrist in each group, cases 11–20 to the second psychiatrist, and so on. This ensured that each group had two psychiatrists reviewing the same set of 10 cases, and across the three groups, a total of six psychiatrists evaluated each set of 10 cases. This design minimized the potential influence of individual differences on the results. During the expert evaluation phase, the six responses from each group were randomized and presented to experts in a blinded manner to ensure the objectivity and reliability of the assessment process.

## Ethics approval

This study adhered to the principles outlined in the Declaration of Helsinki. Informed consent was obtained from each psychiatrist before their participation. Only retrospective clinical data was used and had been fully de-identified prior to access, including removal of all direct identifiers (e.g., names, addresses, contact information) and manual review of free-text content to eliminate any potentially re-identifiable information. The study protocol received approval from the Ethics Committee of Beijing Anding Hospital, Capital Medical University. The individual data were completely anonymous, making the study exempt from requiring informed consent.

## Data and code availability

All the data and code used in this study are accessible at https://github.com/wangrx33/PsychBench. To promote transparency and collaboration within the research community, we have made the full benchmark dataset freely available to the research community for academic use. Researchers can directly download the dataset from our GitHub repository. The dataset is provided in Chinese by default, reflecting its origin in real-world psychiatric clinical settings. A translated English version of the dataset is available upon request, should it be needed for cross-linguistic research or replication purposes.

# Supplementary information

Supplementary Table. S1 | Prompts and examples of input patient information for the 6 tasks in PsychBench.

Supplementary Table. S2 | The example outputs of evaluated models on a random case in terms of 5 clinical tasks.

Supplementary Table. S3 | Definition and examples of error categories in Task1 Clinical text understanding and generation: Clinical text understanding and generation.

Supplementary Table. S4 | Definition and examples of error categories in Task2: Principal diagnosis.

Supplementary Table. S5 | Definition and examples of error categories in Task3: Differential Analysis.

Supplementary Table. S6 | Definition and examples of error categories in Task4: Medication recommendation.

Supplementary Table. S7 | Definition and examples of error categories in Task5: Long-term course management.

Supplementary Table. S8 | The example answers given by each group in reader study.

Supplementary Table. S9 | The example of junior group misclassifying depressive episodes as recurrent depressive disorder.

Supplementary Table. S10 | Example answer given by LLM for differential diagnosis.

Supplementary Table. S11 | The example of LLM assisting medication suggestion.

Supplementary Table. S12 | Reference answers of reader study.

Supplementary Table. S13 | The example of LLMs tended to recommend drugs that had appeared in the medical records.

Supplementary Table. S14 | Overview of quantitative metrics and their clinical interpretations in PsychBench.

Supplementary Table. S15 | Chain of thought prompts used for primary diagnosis and medication recommendation tasks.

Supplementary Table. S16 | The detailed quantitative results of evaluated models on PsychBench in terms of metrics (1-shot).

Supplementary Table. S17 | The detailed quantitative results of evaluated models on PsychBench in terms of metrics (0-shot).

## Acknowledgements

The clinical data is collected from Beijing Anding Hospital, Capital Medical University. We would like to acknowledge the expert review committee at Beijing Anding Hospital for their clinical data audit. Thanks are likewise given to the clinicians of varying seniority at Beijing Anding Hospital who participated in the reader study. This study was funded by the National Key Research and Development Program of China (Grant No. 2022YFB4702702).

## Author contributions

S.L., R.W., L.Z., and X.Z. contributed equally to this study. R.W. and S.L. designed the pipeline of the study, preprocessed data, ran experiments, designed reader study, analyzed results, created figures and wrote the manuscript. All authors reviewed the manuscript and provided meaningful feedbacks. X.Z. collected clinical and analyzed the results. L.Z. and R.Y. organized the reader study and participated in the reader study as specialist psychiatrist. L.Z. provided the clinical guidance of the whole study and participated in the reader study as specialist psychiatrist. X.Z. and F.W. processed data and ran experiments. Z.Y. provided technical advice and assisted in quantitative evaluation. G.W. and C.J. guided the whole study.

## Competing interests

The authors declare no competing interests.

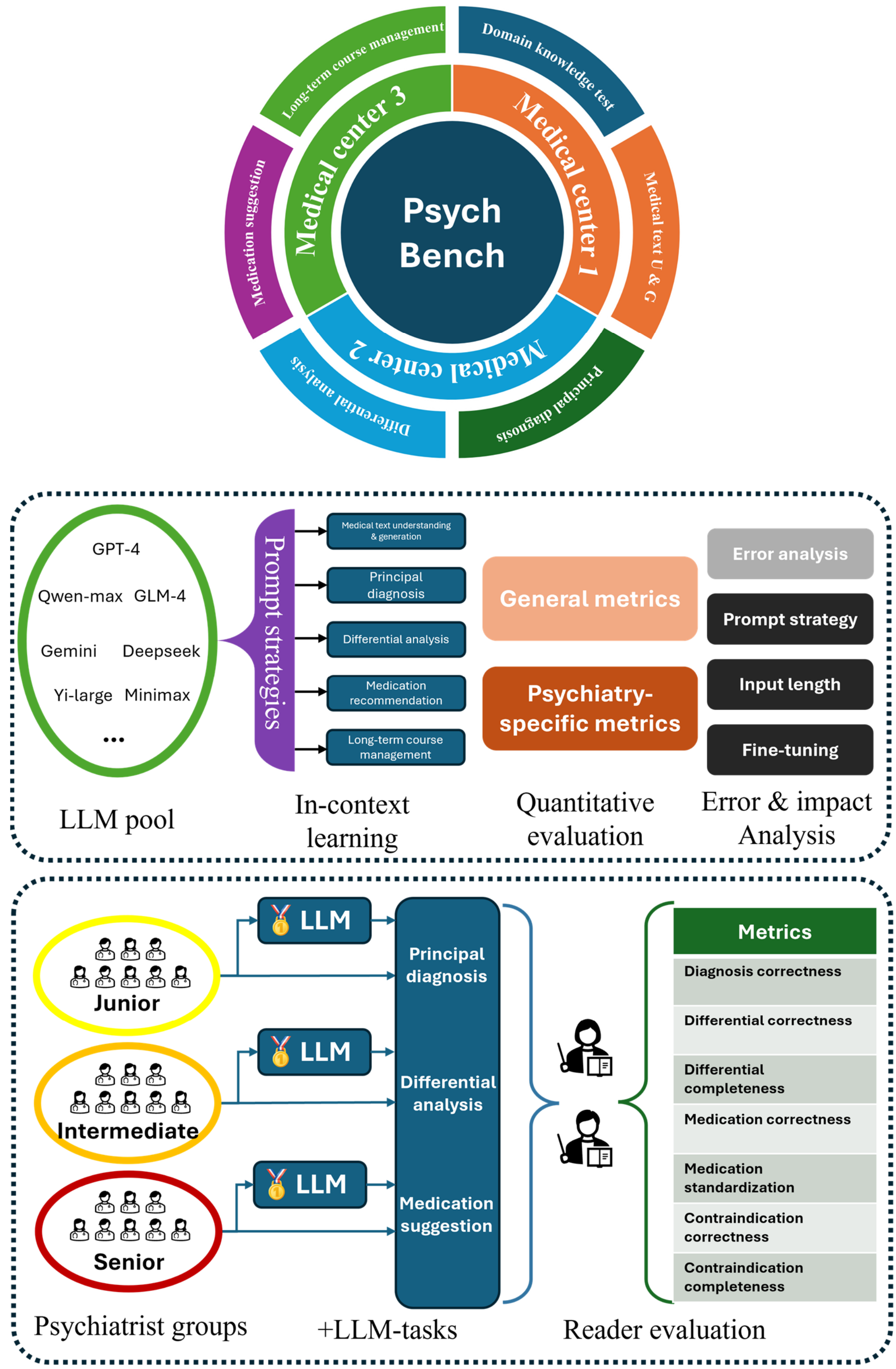

**Fig. 1 | Overview of the framework in this study.** The proposed PsychBench is composed of a dataset and an evaluation framework. The dataset comprises 300 real patient cases collected from three specialized psychiatric medical centers. The evaluation framework consists of five specifically designed psychiatric clinical tasks and corresponding quantitative metrics tailored for each task. The tasks include clinical text understanding and generation, principal diagnosis, differential analysis, medication recommendation, and long-term course management. In this study, we first quantitatively evaluated 16 existing LLMs using PsychBench. We also performed error analysis and assessed the impact of prompt

strategies, input length, and domain-specific fine-tuning on model performance. We then conducted a clinical reader study to further evaluate the effectiveness of LLMs in assisting psychiatrists with different levels of experience. Sixty psychiatrists with varying levels of work experience were recruited to accomplish specific tasks in PsychBench with and without the assistance of LLM respectively. Two specialist psychiatrists then scored the answers given by different groups based on predefined evaluation criteria.

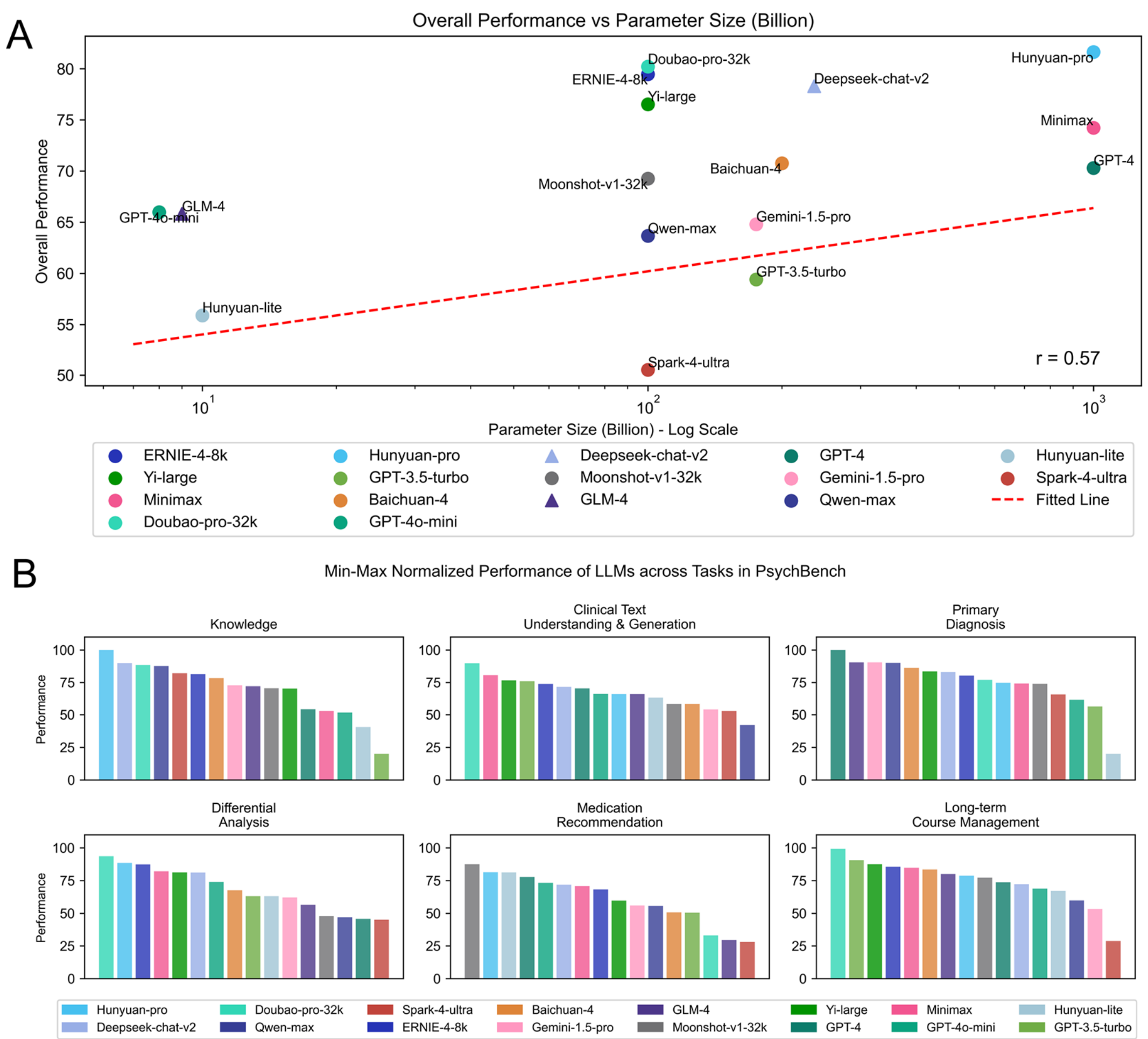

**Fig. 2 | The performance of evaluated LLMs across six tasks in PsychBench. A,** Scatter plot illustrating the relationship between model size (in billions of parameters) and overall performance across six psychiatric tasks. The overall performance score demonstrates a moderate positive correlation with model size (Pearson's r = 0.57), suggesting that larger models tend to exhibit stronger capabilities in the psychiatric care domain. **B,** Task-specific performance of each model based on min-max normalization. For each task, all evaluation metrics were normalized and averaged to derive a composite performance score depicted in panel B. These composite scores were then averaged across all six tasks to generate the overall benchmark score for each model depicted in panel A. The absolute scores of all models on each specific metric are presented in **Extended Data Fig. 1.**

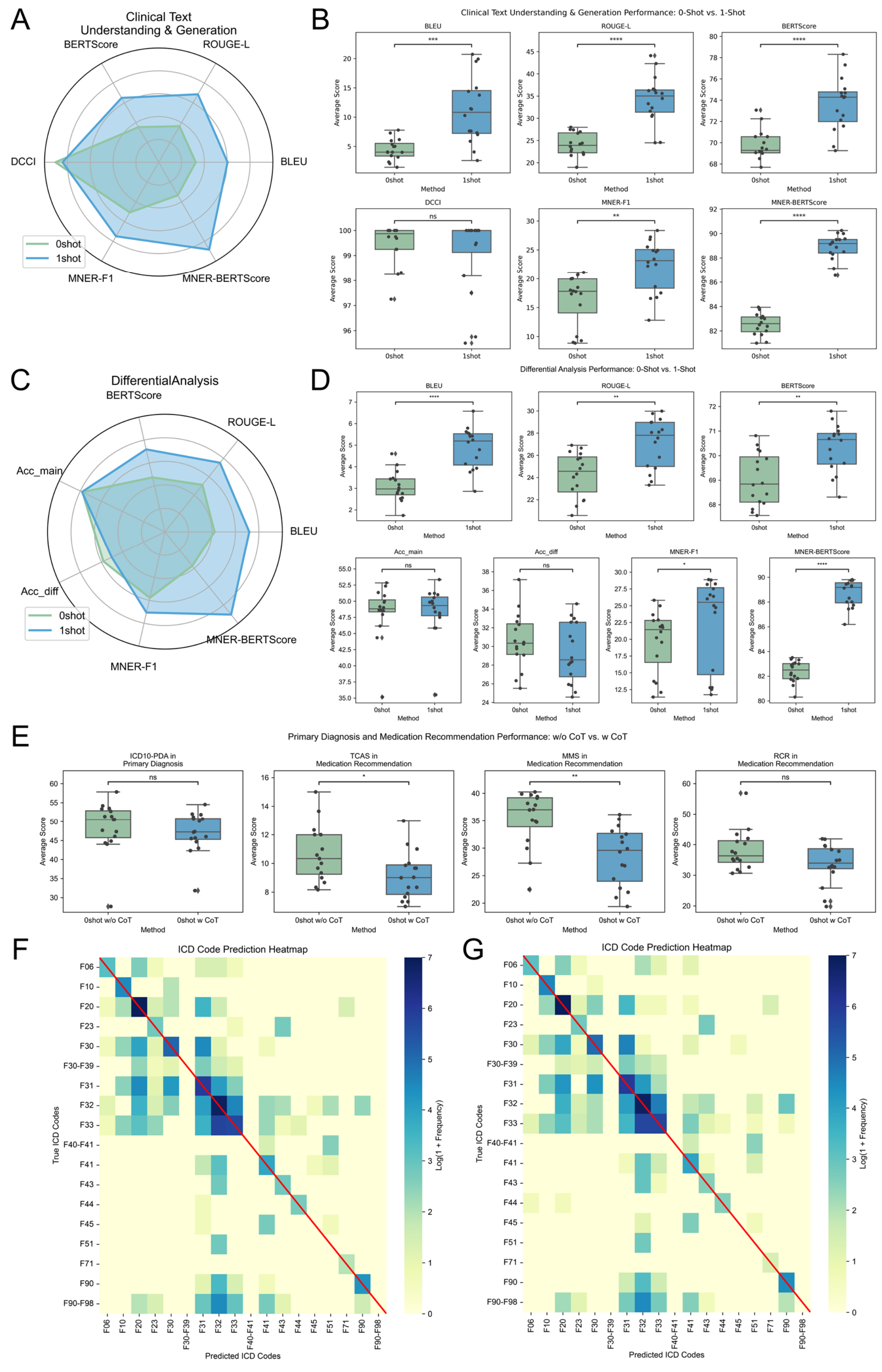

**Fig. 3 | Identifying the influence of prompt strategies: few-shot learning and chain-of-thought (CoT). A-B,** 1-shot prompting notably improved performance on the clinical text understanding and generation task, highlighting LLMs' ability to learn psychiatric documentation styles from minimal examples. **A,** The performance of models with 0-shot prompt and 1-shot prompt on task1 across

*BERTScore*, *ROUGE-L*, *BLEU*, completeness, and accuracy. **B**, The boxplots comparison between models with 0-shot prompt and 1-shot prompt on each metric in task1. **C-D,** in Differential Analysis Task, 1-shot prompting did not improve differential diagnosis accuracy, suggesting a risk of exemplar-induced bias in reasoning tasks. **C**, The performance of models with 0-shot prompt and 1-shot prompt on task3 across *BERTScore*, *ROUGE-L*, *BLEU*, completeness, accuracy of principal diagnosis, and accuracy of differential diagnosis. **D-F,** CoT promptings lead to performance declines on Primary Diagnosis Task and Medication Recommendation Task, indicating that simulated reasoning steps do not translate to clinically appropriate decisions in complex psychiatric scenarios. **D**, The boxplots comparison between models with 0-shot prompt and 1-shot prompt on each metric in task3. **E**, The boxplots comparison between models without CoT prompt (0-shot) and with CoT prompt (0-shot) on each metric in task2 and task4. **F**, The heatmap of the predicted ICD codes given by models without CoT prompt against the reference ICD codes in Primary Diagnosis Task. **G**, The heatmap of the predicted ICD codes given by models with CoT prompt against the reference ICD codes in Primary Diagnosis Task.

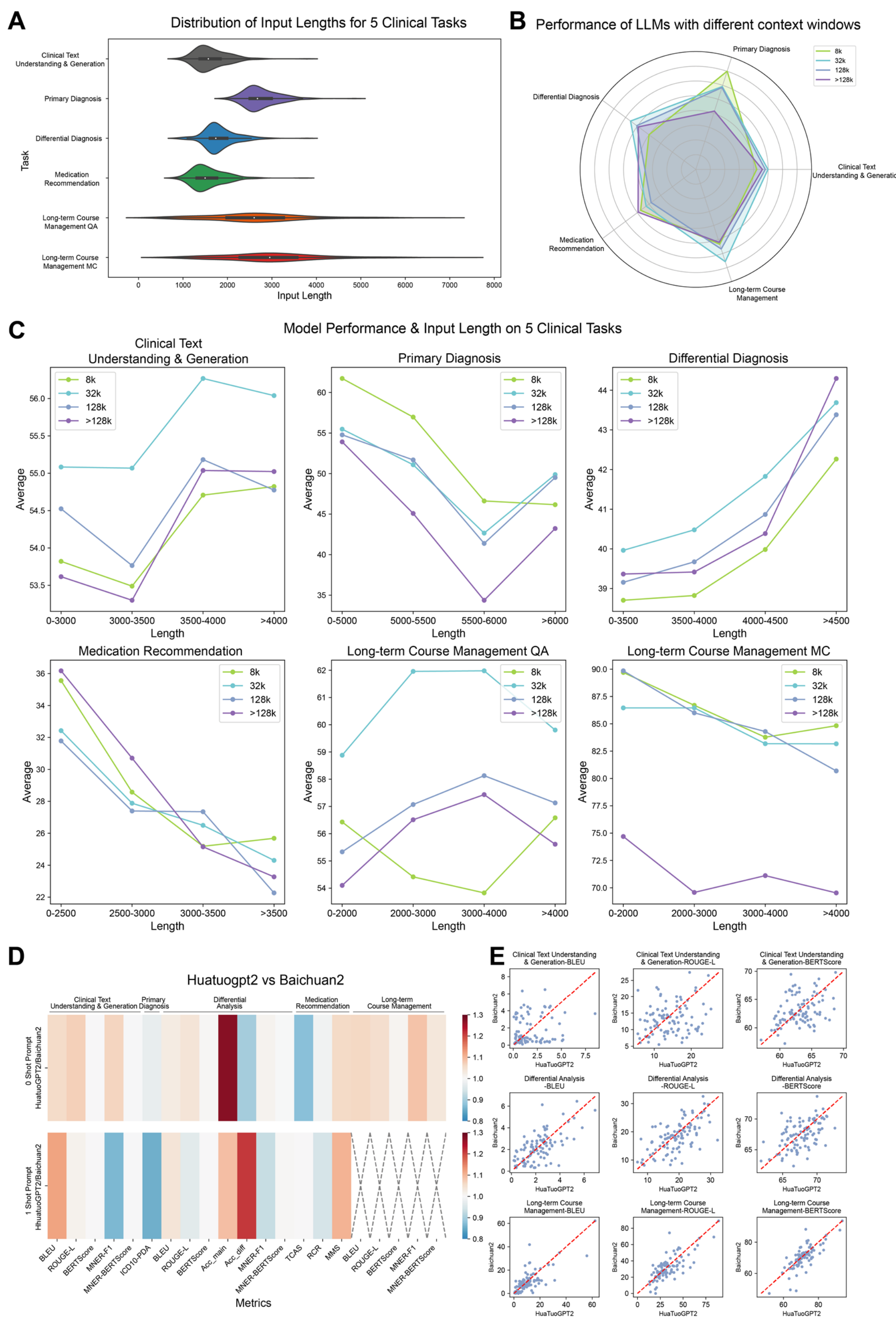

**Fig. 4 | Identifying the influence of the input context length and the medicine-oriented fine-tuning.** **A,** the distribution of the input context lengths across 5 tasks in PsychBench. The unit of length is Chinese words. **B,** The performance of LLMs with different lengths of context windows across five tasks. Since GPT-3.5-turbo has a context length of $16k$, it was not included in this analysis. **C,** How model performance varies with input context length across five tasks. After extending the context window, LLMs do not necessarily "understand" the content better. **D-E,** Performance comparison between medicine-oriented fine-tuned model HuatuoGPT2 and universal model Baichuan2 across each metric of

5 clinical tasks. The medical fine-tuned HuatuoGPT2 exhibits nuanced improvements or comparable performance relative to the general-purpose Baichuan2

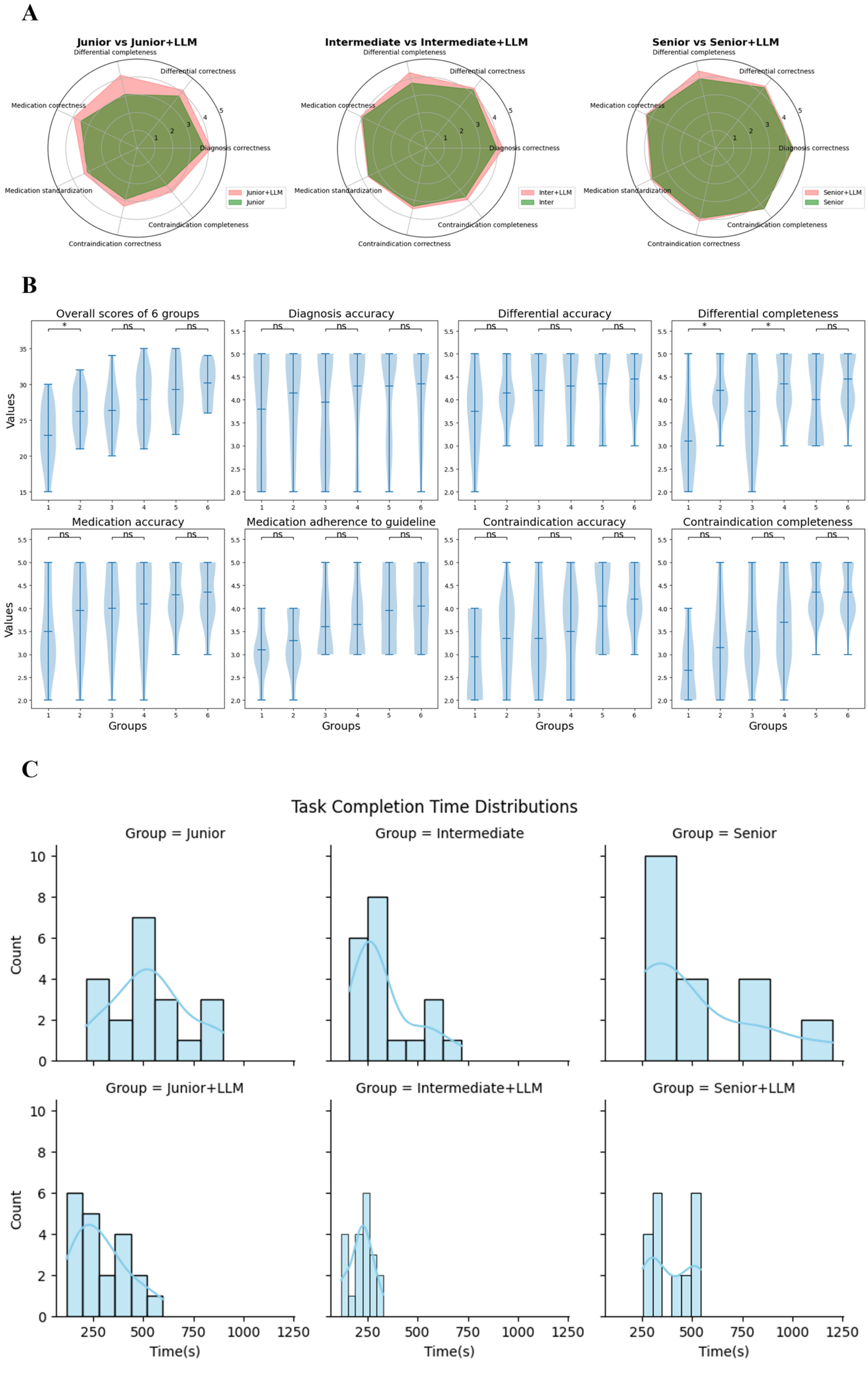

**Fig. 5 | Clinical reader study reveals that LLM assistance improves diagnostic performance and efficiency, particularly for junior psychiatrists. A,** The specialist evaluation of six groups (junior, junior+LLM, intermediate, intermediate+LLM, senior, senior+LLM) across Diagnostic correctness, Differential correctness, Differential completeness, Medication correctness, Medication standardization,

Contraindicated correctness, and Contraindicated completeness. The overall scores are indicated as areas of each radar map. **B,** The comparison of overall scores and scores of each evaluation dimension of six groups. Group 1 to 6 represents group junior, group junior + LLM, group intermediate, group intermediate + LLM, group senior, and group senior + LLM, respectively. '*' indicates a statistically significant difference between the two groups (p-value less than 0.05), while 'ns' indicates no statistically significant difference between the two groups (p-value greater than 0.05). **C,** The distribution of time taken for each group to complete the three clinical tasks—diagnosis, differential analysis, and medication recommendation. LLMs notably reduced task completion time in junior and intermediate groups.

**Extended Data Table. 1 | The Statistics of PsychBench dataset.**

| | **Center 1** | **Center 2** | **Center 3** |
|---|---|---|---|
| **Age, M[IQR]** | 35.0 [25.75, 53.0] | 38.5 [23.0, 52.25] | 26.5 [15.0, 49.0] |
| **Gender, N** | | | |
| - Male | 34 | 13 | 40 |
| - Female | 66 | 87 | 60 |
| **Marriage, N** | | | |
| - Married | 51 | 52 | 41 |
| - Single | 31 | 38 | 55 |
| - Divorced or widowed | 18 | 10 | 4 |
| **Career, N** | | | |
| - Student | 12 | - | 44 |
| - Employed | 46 | - | 42 |
| - Unemployed | 33 | - | 10 |
| - Retired | 9 | - | 4 |
| **Ethnic group, N** | | | |
| - Han | 77 | - | 38 |
| - Man | 12 | - | 0 |
| - Hui | 11 | - | 2 |
| - Zang | 0 | - | 3 |
| - Bai | 0 | - | 32 |
| - Yi | 0 | - | 15 |
| - Other | 0 | - | 10 |
| **Family history, N** | | | |
| - **Yes** | 43 | 30 | 13 |
| - **No** | 57 | 70 | 87 |
| **Duration of illness, M[IQR]** | 4.0 [0.42, 10.0] | 4.0 [2.0, 15.0] | 2.0 [1.0, 5.0] |
| **Principal diagnosis (ICD-10)** | | | |
| - F10.x | 0 | 0 | 6 |
| - F20.x | 20 | 45 | 7 |
| - F30.x | 20 | 2 | 0 |
| - F31.x | 20 | 15 | 6 |
| - F32.x | 20 | 23 | 33 |
| - F33.x | 20 | 15 | 11 |
| - F90.x | 0 | 0 | 8 |
| - F98.x | 0 | 0 | 7 |
| - Others | 0 | 0 | 22 |

**Extended Data Table. 2 | The statistics of evaluated models.**

| Model name | Parameters | Context length | Type |
|---|---|---|---|
| **Baichuan4** | X00B+ | 32k | Close-Source General Domain |
| **Deepseek** | 236B | 128k | Open-Source General Domain |
| **Doubao-pro-32k** | 100B+(MoE) | 32k | Close-Source General Domain |
| **ERNIE-4-8k** | 100B+ | 8k | Close-Source General Domain |
| **Gemini-1.5-pro** | 175B (MoE) | 2M | Close-Source General Domain |
| **GLM-4** | 9B | 128k | Open-Source General Domain |
| **Hunyuan-lite** | 10B+ (MoE) | 256k | Close-Source General Domain |
| **Hunyuan-pro** | 1T+ (MoE) | 32k | Close-Source General Domain |
| **Minimax** | 1T+ | 245k | Close-Source General Domain |
| **Moonshot-v1-32k** | 100B+ | 32k | Close-Source General Domain |
| **Qwen-max** | 100B+ | 8k | Close-Source General Domain |
| **Spark-4ultra** | 100B+ | 8k | Close-Source General Domain |
| **Yi-large** | 100B+ | 32k | Close-Source General Domain |
| **GPT-3.5-turbo** | 175B | 16k | Close-Source General Domain |
| **GPT-4o-mini** | Est. 8B | 128k | Close-Source General Domain |
| **GPT-4** | 1T+ (MoE) | 8k | Close-Source General Domain |
| **Baichuan2-7b** | 7B | 4k | Open-Source Medical Domain |
| **HuatuoGPT2-7b** | 7B | 4k | Open-Source Medical Domain |

**Extended Data Table. 3 | The leaderboard of LLMs on PsychBench.**

| Rank | Model | Task0 Knowledge | Task1 Clinical text understanding and generation | Task2 Principal diagnosis | Task3 Differential Analysis | Task4 Medication recommendation | Task5 Long-term course management | Overall |
|---|---|---|---|---|---|---|---|---|
| 1 | Hunyuan-pro | 100.00 | 66.18 | 74.84 | 88.55 | 81.43 | 78.78 | 81.63 |
| 2 | Doubao-pro-32k | 88.40 | 89.81 | 77.01 | 93.66 | 33.08 | 99.29 | 80.21 |
| 3 | ERNIE-4-8k | 81.43 | 73.84 | 80.27 | 87.38 | 68.30 | 85.64 | 79.48 |
| 4 | Deepseek | 89.98 | 71.60 | 82.99 | 81.13 | 71.88 | 72.25 | 78.30 |
| 5 | Yi-large | 70.37 | 76.71 | 83.54 | 81.31 | 59.74 | 87.53 | 76.53 |
| 6 | Minimax | 52.84 | 80.61 | 74.30 | 82.13 | 70.75 | 84.72 | 74.22 |
| 7 | Baichuan4 | 78.37 | 58.22 | 86.24 | 67.56 | 50.73 | 83.47 | 70.77 |
| 8 | GPT-4 | 54.12 | 70.47 | 100.00 | 45.72 | 77.76 | 73.70 | 70.30 |
| 9 | Moonshot-v1-32k | 70.62 | 58.25 | 74.01 | 47.97 | 87.48 | 77.24 | 69.26 |
| 10 | GPT-4o-mini | 51.65 | 66.30 | 61.81 | 73.94 | 73.31 | 68.85 | 65.98 |
| 11 | GLM-4 | 72.20 | 66.15 | 90.49 | 56.47 | 29.53 | 80.06 | 65.82 |
| 12 | Gemini-1.5-pro | 72.94 | 53.95 | 90.49 | 62.15 | 56.02 | 53.29 | 64.81 |
| 13 | Qwen-max | 87.65 | 41.92 | 90.08 | 46.96 | 55.56 | 59.87 | 63.67 |
| 14 | GPT-3.5-Turbo | 20.00 | 75.99 | 56.30 | 63.15 | 50.46 | 90.59 | 59.41 |
| 15 | Hunyuan-lite | 40.49 | 63.41 | 20.00 | 63.05 | 81.30 | 67.12 | 55.89 |
| 16 | Spark-4-Ultra | 82.22 | 52.92 | 65.88 | 45.07 | 28.10 | 28.84 | 50.51 |

**Extended Data Table. 4 | The evaluation criteria designed for reader study.**

| **The Evaluation Criteria: 7 dimensions for clinical performance assessment** | | | |
|---|---|---|---|
| | **Evaluation dimension** | **Example scenarios** | **Rating (1-5)** |
| **1** | **Diagnosis accuracy** | Assess whether the clinician accurately identifies and establishes the primary diagnosis using appropriate tools and methods, avoiding misdiagnosis. | ☐ 5 - Excellent<br>☐ 4 - Good<br>☐ 3 - Fair<br>☐ 2 - Needs Improvement<br>☐ 1 - Poor |
| | | **Scenario 1**: The clinician accurately assesses and confirms the primary psychiatric diagnosis based on symptoms, history, and clinical examination. | |
| | | **Scenario 2**: The clinician reviews prior diagnoses to prevent treatment delays caused by initial diagnostic errors. | |
| **2** | **Differential Accuracy** | Evaluate whether the clinician accurately rules out potential misdiagnoses, ensuring the diagnosis aligns with the patient's clinical presentation. | ☐ 5 - Excellent<br>☐ 4 - Good<br>☐ 3 - Fair<br>☐ 2 - Needs Improvement<br>☐ 1 - Poor |
| | | **Scenario 1**: The clinician confirms the primary diagnosis through further examination, ruling out any misdiagnoses. | |
| | | **Scenario 2**: The clinician uses clinical evidence to eliminate possible misdiagnoses. | |
| **3** | **Differential Completeness** | Assess whether the clinician comprehensively considers alternative diagnoses similar to the primary diagnosis, covering all relevant possibilities. | ☐ 5 - Excellent<br>☐ 4 - Good<br>☐ 3 - Fair<br>☐ 2 - Needs Improvement<br>☐ 1 - Poor |
| | | **Scenario 1**: The clinician lists potential differential diagnoses, covering all major conditions for a thorough assessment. | |
| | | **Scenario 2**: The clinician conducts a comprehensive analysis based on history and symptoms, ruling out all relevant psychiatric conditions. | |
| **4** | **Medication Accuracy** | Evaluate whether the clinician's medication recommendations align with the diagnosis and the patient's specific needs, ensuring the appropriateness of drug selection and dosage. | ☐ 5 - Excellent<br>☐ 4 - Good<br>☐ 3 - Fair<br>☐ 2 - Needs Improvement<br>☐ 1 - Poor |
| | | **Scenario 1**: The medication was in accordance with the medication specifications, and there were no basic medication errors. | |
| | | **Scenario 2**: The medication was consistent with the patient's symptoms. | |
| **5** | **Medication Adherence to Guidelines** | Assess whether the clinician's medication recommendations follow clinical guidelines and standards, avoiding inappropriate practices. | ☐ 5 - Excellent<br>☐ 4 - Good<br>☐ 3 - Fair<br>☐ 2 - Needs Improvement<br>☐ 1 - Poor |
| | | **Scenario 1**: The clinician prescribes according to the latest clinical guidelines, with no inappropriate medication practices. | |
| | | **Scenario 2**: The clinician consults clinical guidelines before prescribing to ensure an evidence-based decision. | |
| **6** | **Contraindication Accuracy** | Verify whether the clinician accurately identifies and avoids contraindicated medications to ensure safe prescribing. | ☐ 5 - Excellent<br>☐ 4 - Good<br>☐ 3 - Fair<br>☐ 2 - Needs Improvement<br>☐ 1 - Poor |

| | | **Scenario 1**: The clinician identifies contraindications in the patient's profile and selects alternative medication. | |
|---|---|---|---|
| | | **Scenario 2**: The clinician thoroughly reviews the patient's history to ensure no contraindicated medications are prescribed. | |
| **7** | **Contraindication Completeness** | Assess whether the clinician thoroughly considers the patient's allergy history, past medical history, and potential drug interactions to avoid contraindications. | ☐ 5 - Excellent<br>☐ 4 - Good<br>☐ 3 - Fair<br>☐ 2 - Needs Improvement<br>☐ 1 - Poor |
| | | **Scenario 1**: The clinician gathers a comprehensive medication and medical history from the patient to avoid drug interaction risks. | |
| | | **Scenario 2**: The clinician assesses contraindications based on past medical and allergy history, avoiding all potential contraindicated medications. | |

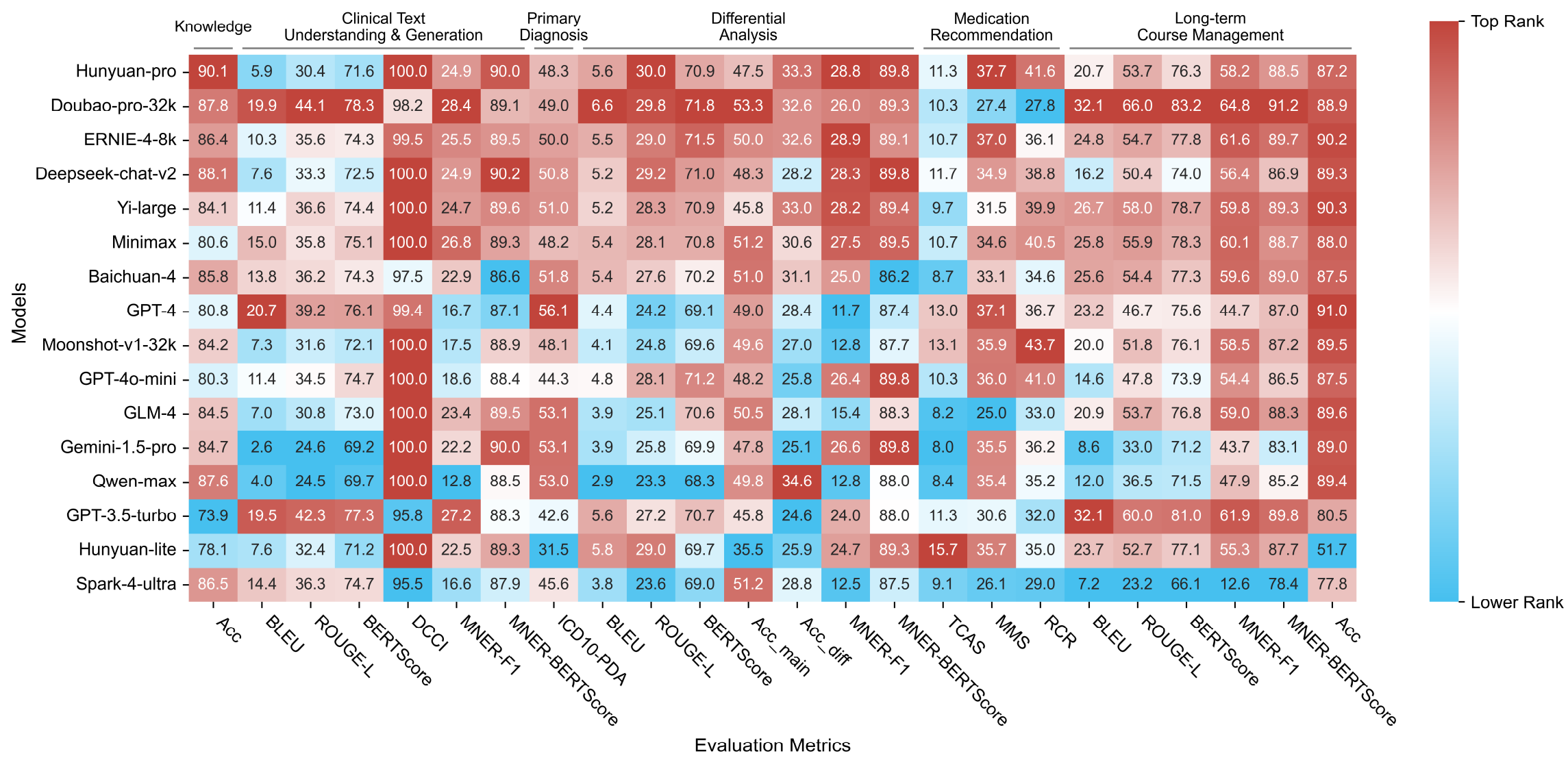

**Extended Data Fig. 1 | Heatmap of absolute scores for all individual evaluation metrics across tasks and models.** This heatmap displays the absolute scores of each model on every individual evaluation metric within the six psychiatric tasks. Color intensity ranges from blue to red, representing the relative ranking of models from lowest to highest on each specific metric. This visualization highlights performance disparities across models at a granular level and complements the aggregated results shown in Fig. 2.

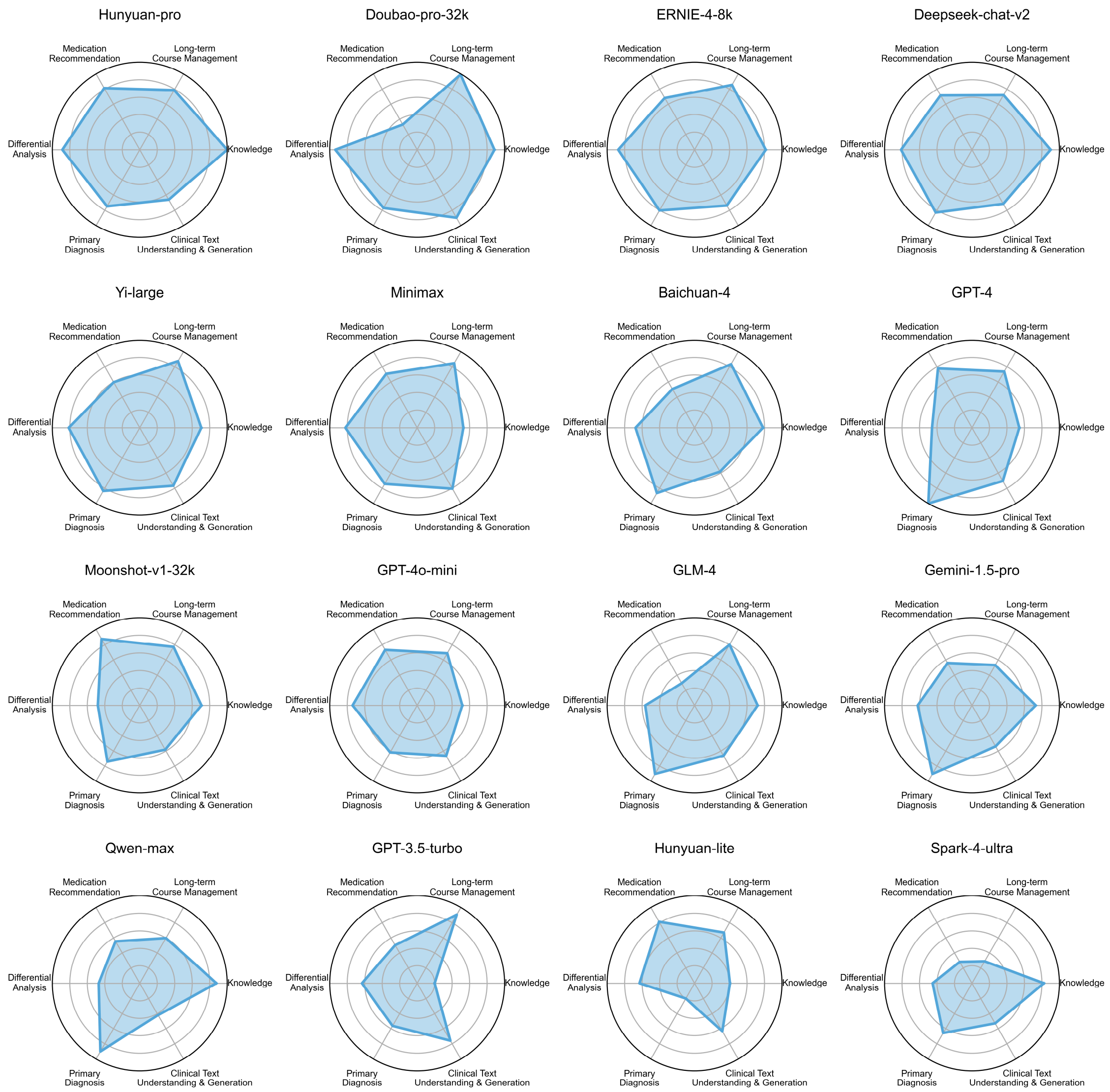

**Extended Data Fig. 2 | Radar plots of the normalized quantitative results of evaluated LLMs across six tasks.** We applied min-max normalization to revalue each evaluation metric for every task and then calculated the mean of all metrics for each task as the overall performance indicator of the corresponding task. Subsequently, we computed the mean of the overall performance indicators across the six tasks to serve as the comprehensive evaluation metric for the large models in the field of psychiatric care as indicated as the area of each radar map. The radar maps are arranged from left to right and top to bottom in descending order of the comprehensive evaluation metric, reflecting the overall performance of each large model in the psychiatric care domain.

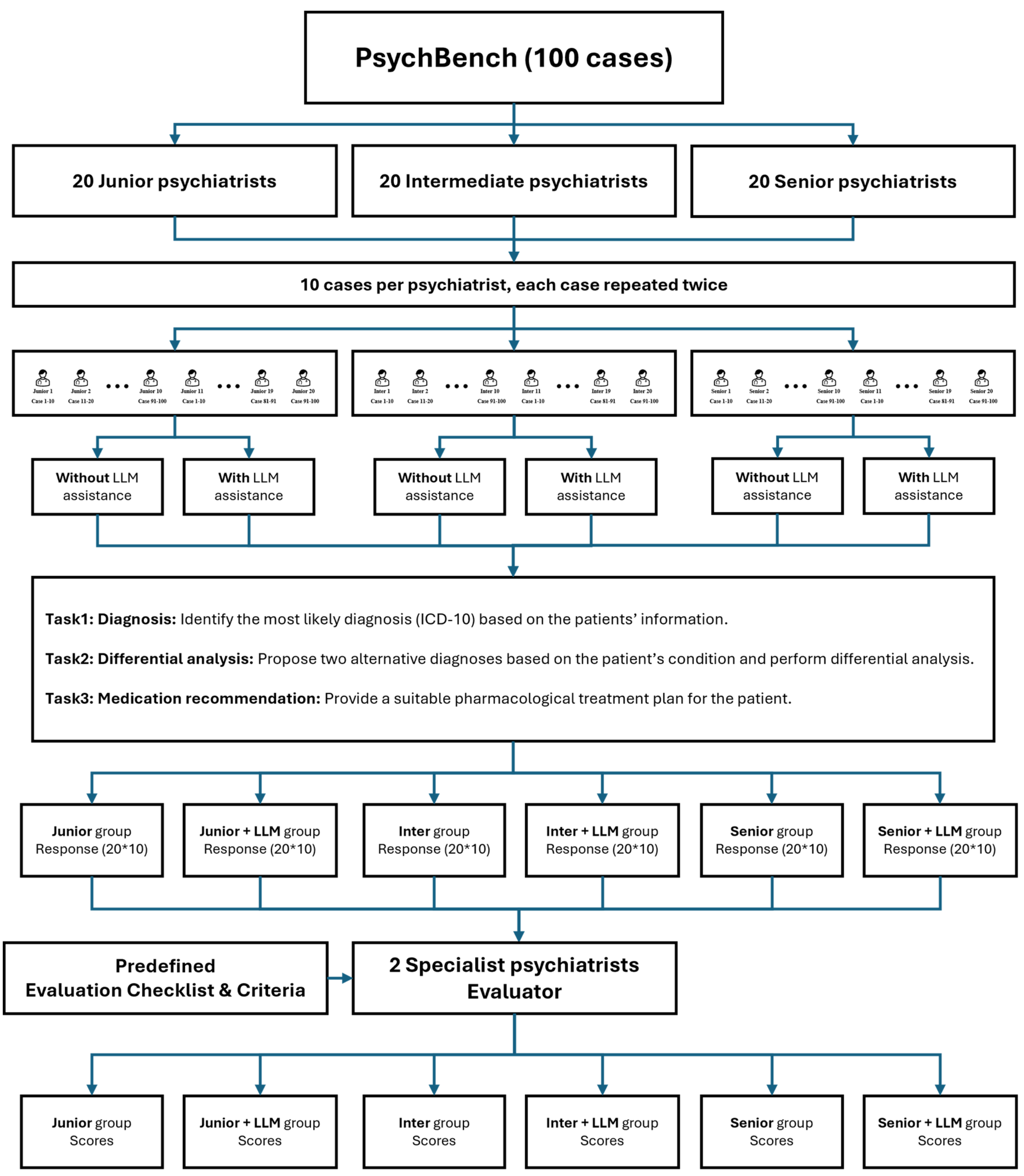

**Extended Data Fig. 3 | Study design of the clinical reader study.** We conducted a reader study with 60 psychiatrists (20 junior, 20 intermediate, 20 senior), each completing clinical tasks—including diagnosis, differential diagnosis, and medication recommendation—on 10 real-world psychiatric cases under two conditions: with and without LLM assistance. This yielded six groups based on experience level and assistance type. Responses were scored by 2 expert raters using a scoring system developed from ICD-10 guidelines and the SaferDx framework (**Extended Data Table 4**), and task completion times were recorded.

| Input | Group1 | Group2 | Group3 | Group4 | Group5 | Group6 |
|---|---|---|---|---|---|---|
| Patient is a male, 58 years old. Starting in early 2020, without any apparent cause, the patient gradually began to experience disordered thinking, excessive rumination, and increased worry. He developed a low mood, reluctance to engage in activities… | **Task1:** F32.301 | **Task1:** F32.301 | **Task1:** F33.301 | **Task1:** F32.301 | **Task1:** F32.301 | **Task1:** F32.301 |
| | **Task2:** Schizophrenia… | **Task2:** Schizophrenia,GAD.. | **Task2:** BD,Schizophrenia… | **Task2:** BD,GAD, DD… | **Task2:** GAD… | **Task2:** GAD,DD … |
| | **Task3:** Duloxetine … | **Task3:** Mirtazapine … | **Task3:** Escitalopram… | **Task3:** Aripiprazole… | **Task3:** Milnacipran… | **Task3:** Aripiprazole… |
| Diagnosis correctness | 5 | 5 | 2 | 5 | 5 | 5 |
| Differential correctness | 3 | 4 | 2 | 4 | 3 | 4 |
| Differential completeness | 3 | 3 | 2 | 3 | 2 | 3 |
| Medication correctness | 5 | 3 | 2 | 3 | 3 | 3 |
| Medication standardization | 5 | 3 | 2 | 3 | 3 | 3 |
| Contraindication correctness | 1 | 4 | 3 | 4 | 3 | 4 |
| Contraindication completeness | 1 | 3 | 2 | 3 | 2 | 3 |

**Extended Data Fig. 4 | The reader study user interface.** Interface used by expert psychiatrists to evaluate participants' responses across diagnostic, differential, and treatment tasks. Ellipses ("…") indicate that some content is not fully displayed.

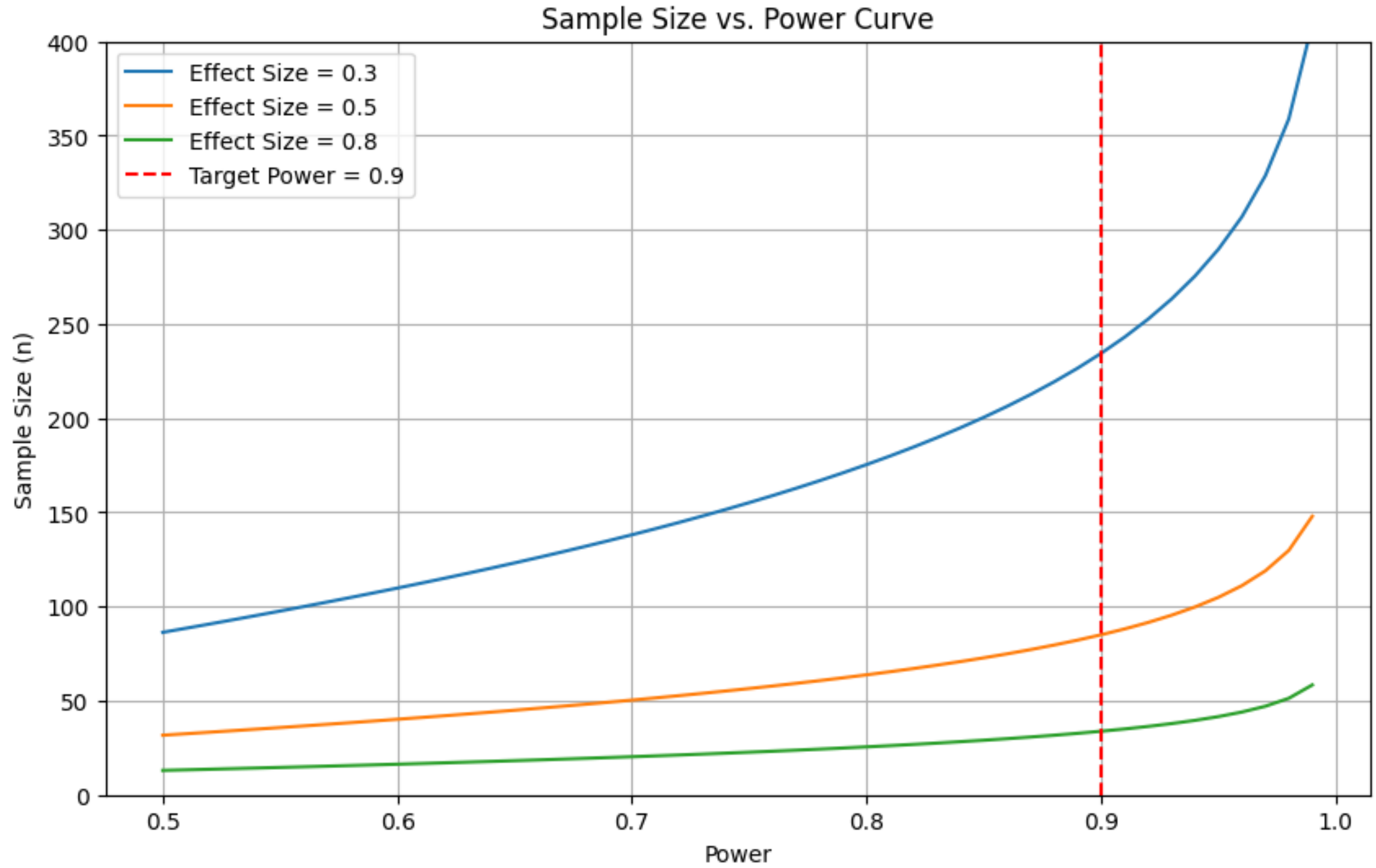

**Extended Data Fig. 5 | Statistical power analysis for sample size determination.** Power analysis used to estimate the minimum required sample size for detecting a medium effect (Cohen's d = 0.5) with 90% power at a significance level of α = 0.05. Based on this analysis, a sample size of at least 86 patients was determined. Ultimately, we collected clinical patient data from each center, totaling 300 cases, with 100 cases from each facility. This ensures sufficient statistical power and enhances the validity of benchmark conclusions.

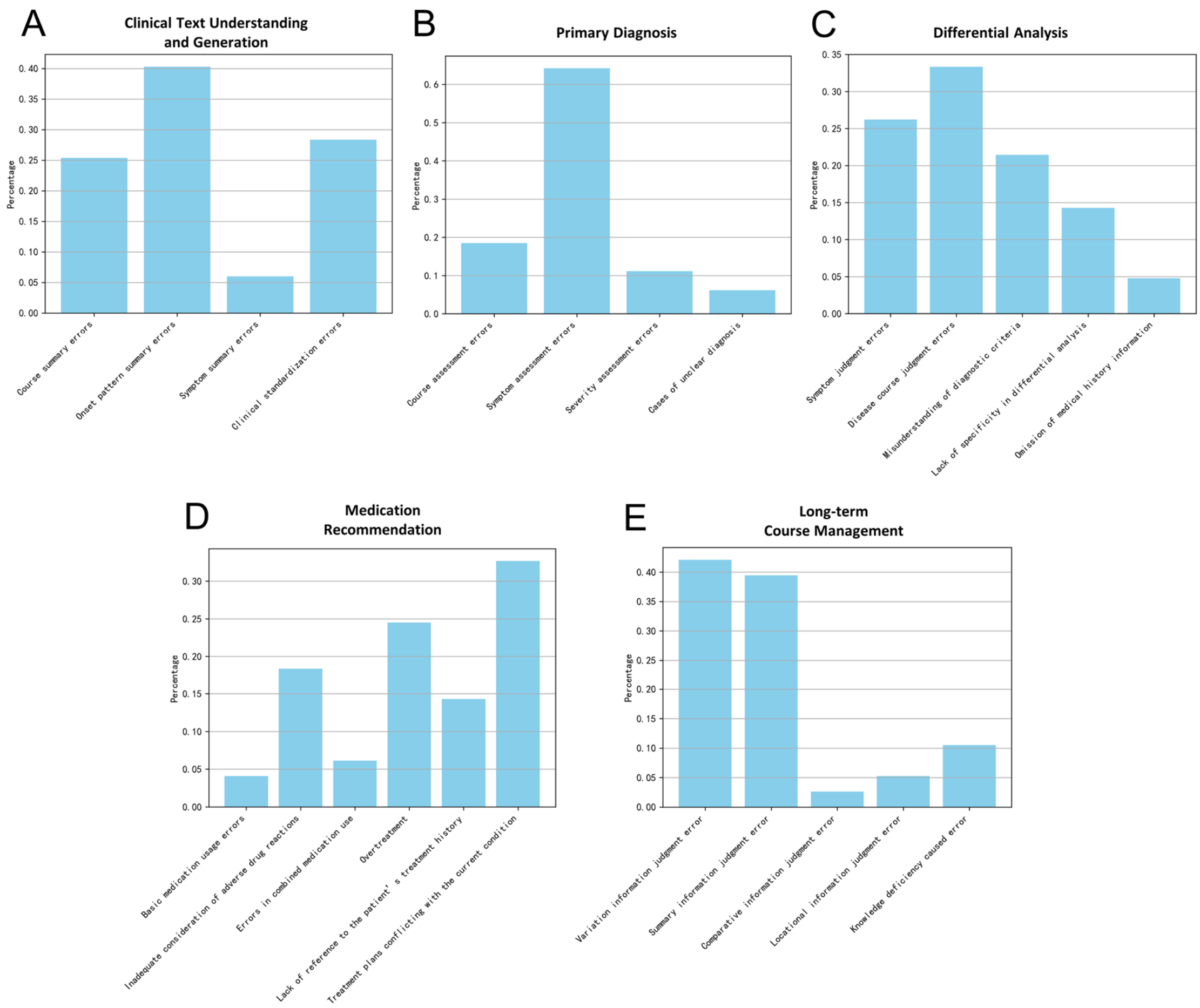

**Extended Data Fig. 6 | Error type distribution for each clinical task in PsychBench.** Each subplot represents one of the five clinical tasks in PsychBench. Bars indicate the proportion of different error types among all error cases within each task. This analysis reveals common failure patterns and informs targeted improvement of LLM performance in psychiatric applications.

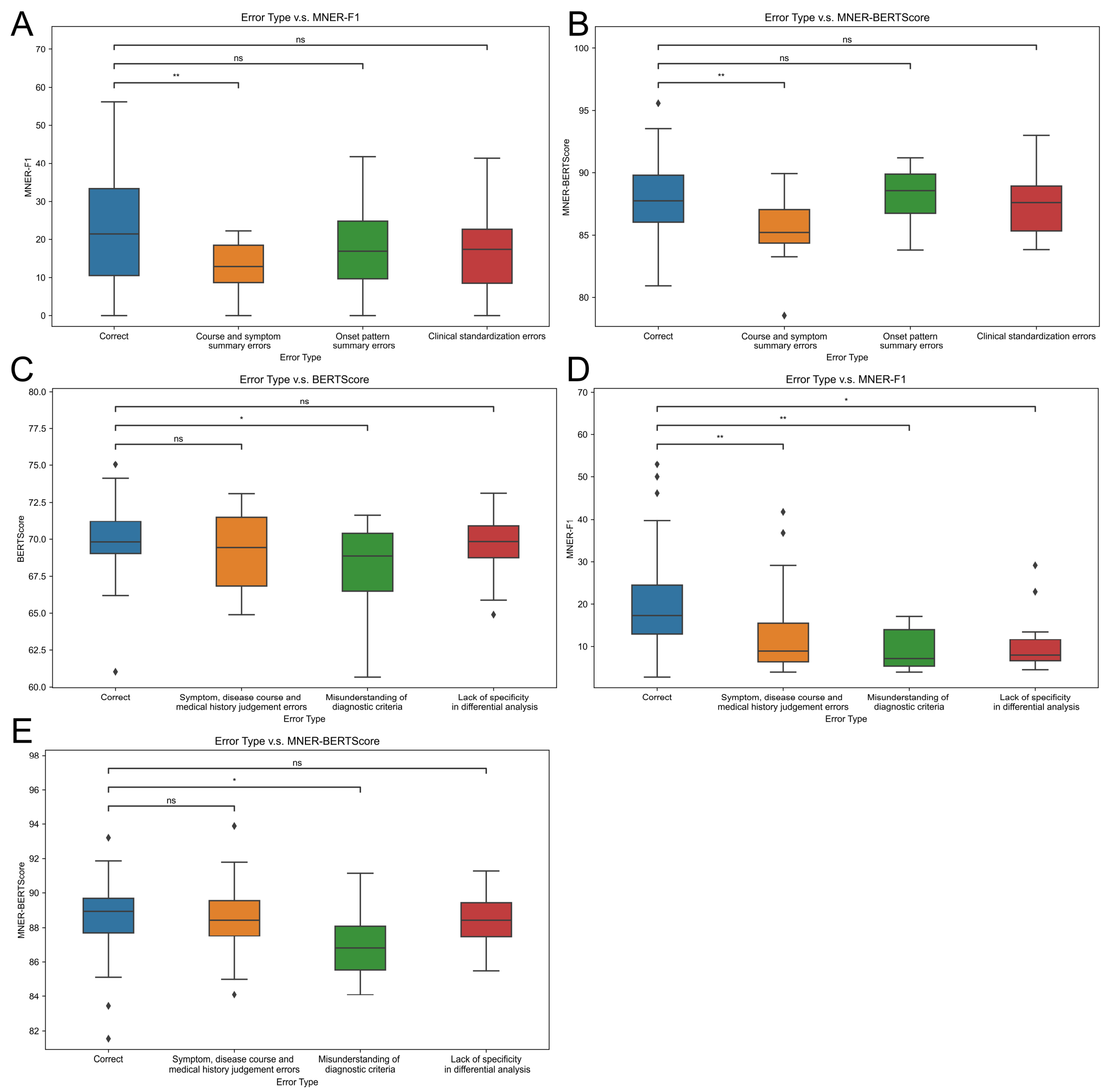

**Extended Data Fig. 7 | The statistics regarding model errors and quantitative metric scores in Clinical Text Understanding and Generation Task (A,B) and Differential Analysis Task (C,D,E). A-B,** *MNER-F1* and *MNER-BERTScore* are significantly higher in cases without course or symptom summary errors in the Clinical Text Understanding and Generation Task (independent t-test p-value < 0.05). **C-E,** In the Differential Analysis Task, the *MNER-F1* score for cases correctly answered by the model is significantly higher than the scores for all other error type groups (independent t-test p-value <0.05). Moreover, *BERTScore* and *MNER-BERTScore* in correct cases are significantly higher than those in error cases due to misinterpreted diagnostic criteria. These observations further justify the design of PsychBench metrics.

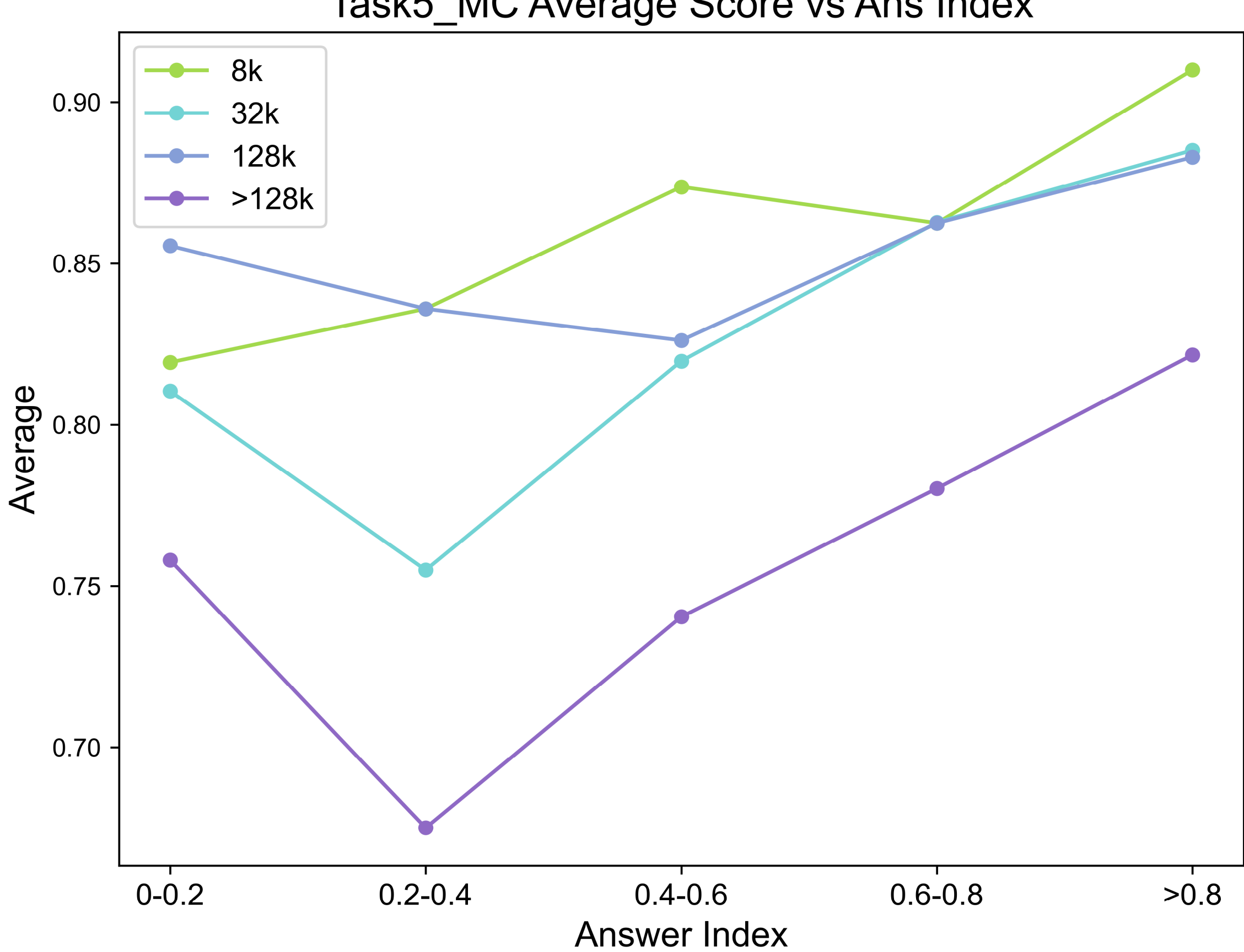

**Extended Data Fig. 8 | The impact of answer position on the model performance on Long-term Course Management Task (Multiple-choice Subtask).** Model accuracy varies with the position of the correct answer in long medical records. For models with ≥32k context length, accuracy drops significantly when answers appear in the middle (0.2–0.4 range).

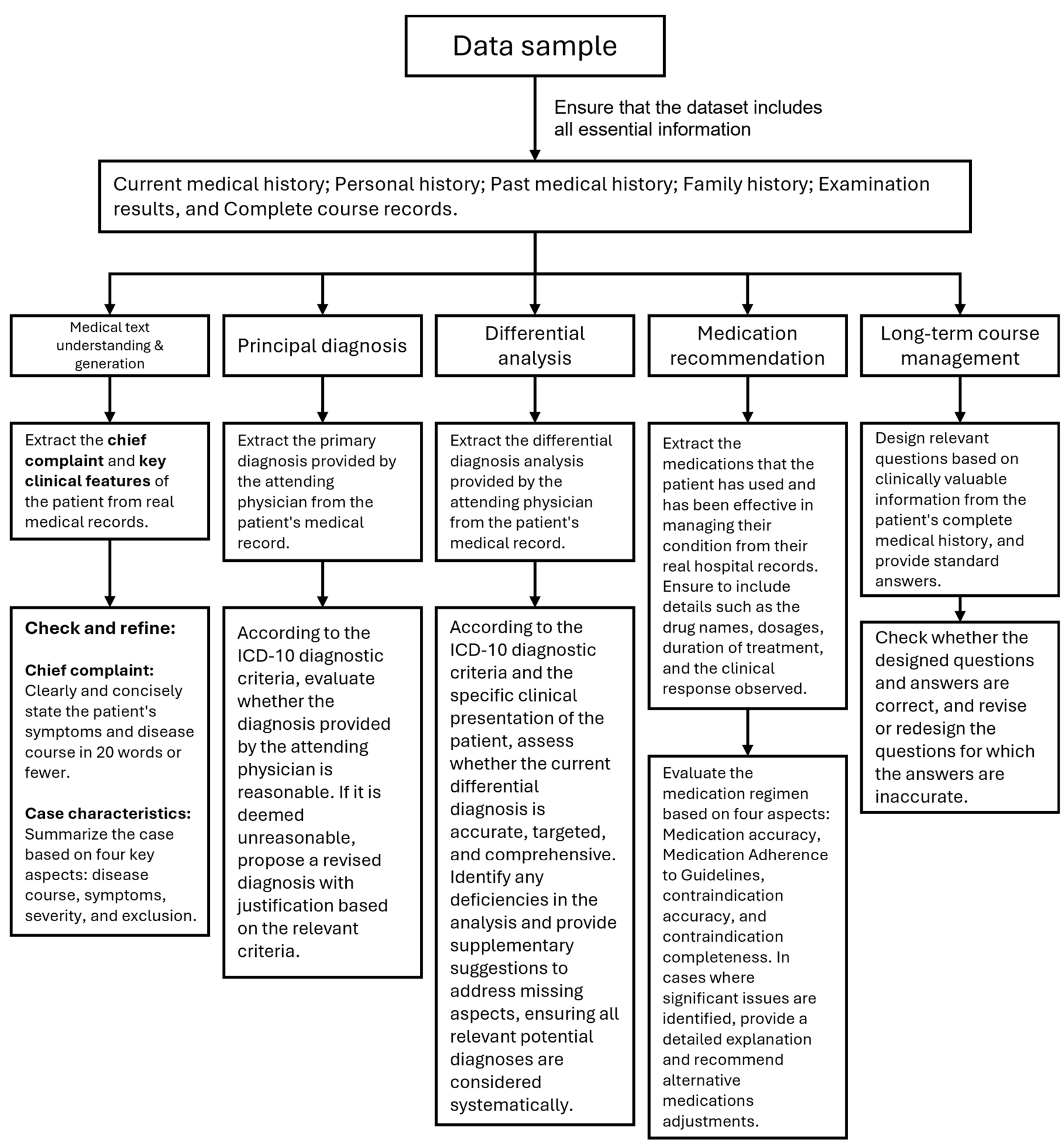

**Extended Data Fig. 9 | The guidelines for data verification and annotation.** This figure outlines the procedures for data verification and annotation, ensuring consistency across institutions and compliance with local ethical and cultural standards. By including data from multiple regions and ethnic groups, the study enhances the representativeness and cross-cultural applicability of PsychBench.